\documentclass[10pt]{article}
\usepackage[utf8]{inputenc}

\usepackage[american]{babel}
\usepackage{natbib} % has a nice set of citation styles and commands
\usepackage{mathtools} % amsmath with fixes and additions
\usepackage{booktabs} % commands to create good-looking tables
\usepackage{tikz} % nice language for creating drawings and diagrams

\usepackage[utf8]{inputenc}
\usepackage{amsmath,mathpazo,mathtools,bm,amsthm}

\usepackage{subcaption}
\usepackage{tablefootnote}
\usepackage{paralist}
\usepackage{pgfplots}
\usetikzlibrary{arrows}
\usetikzlibrary{patterns}
\usepackage{algorithm} 
\usepackage{algorithmic}
\usepackage{url}

\usepackage{multirow}

\usepackage{authblk}

\usepackage{changepage}
\usepackage{dsfont}
\usepackage{tabularx}
\usepackage{amssymb}

\usepackage{stackengine}
\def\delequal{\mathrel{\ensurestackMath{\stackon[1pt]{=}{\scriptstyle\Delta}}}}

\newcommand{\rightleftc}{\circ\!\!-\!\!\circ}
\newcommand{\leftcrighta}{\circ\!\!\!\rightarrow}

\newcommand{\rightc}{-\!\circ}
\newcommand{\rightlefta}{\leftarrow\!\!\!\rightarrow}

\newcommand{\rightclefts}{*\!\!-\!\!\circ}
\newcommand{\leftcrights}{\circ\!\!-\!\!\-*}
\newcommand{\rightalefts}{*\!\!\!\rightarrow}
\newcommand{\leftarights}{\leftarrow\!\!\!*}

\newtheorem{ER-rule}{ER-rule}
\newtheorem{LER-rule}{LER-rule}
\newtheorem{PC-rule}{PC-rule}
\newtheorem{FCI-rule}{FCI-rule}
\newtheorem{FCI-rule8}{FCI-rule}

\newtheorem{Definition}{Definition}%[section]
\newtheorem{Property}{Property}%[section]
\newtheorem{Theorem}{Theorem}%[section]
\newtheorem{Example}{Example}%[section]
\author[1,2]{Charles K. Assaad}
\author[2]{Emilie Devijver}
\author[2]{Eric Gaussier}
\affil[1]{EasyVista}
\affil[2]{Univ. Grenoble Alpes, CNRS, Grenoble INP, LIG}

\title{Entropy-based Discovery of Summary Causal Graphs  in Time Series}

\date{}

\begin{document}
\maketitle

\begin{abstract}
		This study addresses the problem of learning a summary causal graph on time series with potentially different sampling rates. To do so, we first propose a new causal temporal mutual information measure for time series. We then show how this measure relates to an entropy reduction principle that can be seen as a special case of the probability raising principle. We finally combine these two ingredients in PC-like and FCI-like algorithms to construct the summary causal graph. There algorithm are evaluated on several datasets, which shows both their efficacy and efficiency. % of directly learning summary graphs.
		%experiments which shows that directly learning a summary grpah can be both efficient revealed that our approach outperforms previously proposed ones on all the datasets considered.
\end{abstract}

% \textcolor{blue}{
% Assumptions and questions:
% \begin{itemize}
% 	\item Causal Markov Condition
% 	\item Faithfulness
% 	\item Temporal Priority
% 	\item Entropy Reduction
% 	\item Acyclicity in Summary graph 
%   \item self causes are one order markov
%   \item unitary causal relations
% \end{itemize}
% Contribution:
% \begin{itemize}
% 	\item Reduce the amount of permutation test in the search of summary graph 
% 	\item Entropy reduction principle
% 	\item 2 information theoretic measures for time series: CTMI and NCTMI
% 	\item handle different sampling rate
% \end{itemize}
% }

	\section{Introduction}

\label{sec:chapEntropyBased_intro}

Time series arise as soon as observations, from sensors or experiments, for example, are collected over time. They are present in various forms in many different domains, such as healthcare (through, {e.g.}, monitoring systems), Industry 4.0 (through,~{e.g.}, predictive maintenance and industrial monitoring systems), surveillance systems (from images, acoustic signals, seismic waves, etc.), or energy management (through, {e.g.}, energy consumption data), to name but a few. 
We are interested in this study in analyzing {quantitative discrete-time series} to detect the causal relations that exist between them under the assumption of consistency throughout time \citep{Assaad_2022} {which states that causal relationships remain constant in direction throughout time.} %MDPI: The footnote is not allowed, moved into the text, please conform. (Confirmed and rephrased to remove parenthesis) 
Under this assumption, one can replace the infinite full-time causal graph (Figure~\ref{fig:full_vs_summary}a)
%	\begin{Definition}[Full Time Causal Graph]
%		Let $\mathcal{X}$ be a multivariate discrete-time stochastic process and $\mathcal{G} = (V, E)$ the associated \emph{full time causal graph}. The set of vertices in that graph consists of the set of components $\mathcal{X}^1,\ldots, \mathcal{X}^d$ at each time $t \in \mathbb{Z}$. The edges $E$ of the graph are defined as follows: variables $\mathcal X^p_{t-i}$ and $\mathcal X^q_t$ are connected by a lag-specific directed link $\mathcal X^p_{t-i} \rightarrow \mathcal X^q_t$ in $\mathcal{G}$ pointing forward in time if and only if $\mathcal X^p$ causes $\mathcal X^q$ at time $t$ with a time lag of $i> 0$ for $p= q$ and with a time lag of $i\geq 0$ for $p\ne q$.
%	\end{Definition}
by a window causal graph (Figure~\ref{fig:full_vs_summary}b) defined as follows: 
\begin{Definition}[Window causal graph \cite{Assaad_2022}]
	{Let $\mathcal{X}$ be a multivariate discrete-time series and $\mathcal{G} = (V, E)$ the associated \emph{window causal graph} for a window of size $\tau$. The set of vertices in that graph consists of the set of components $\mathcal{X}^1,\ldots, \mathcal{X}^d$ at each time $t , \ldots, t+\tau$. The edges $E$ of the graph are defined as follows: variables $\mathcal X^p_{t-i}$ and $\mathcal X^q_t$ are connected by a lag-specific directed link $\mathcal X^p_{t-i} \rightarrow \mathcal X^q_t$ in $\mathcal{G}$ pointing forward in time if and only if $\mathcal X^p$ causes $\mathcal X^q$ at time $t$ with a time lag of $0\leq i\leq \tau$ for $p\ne q$ and with a time lag of $0<i\leq \tau$ for $p= q$.
	}
\end{Definition}

The window causal graph only covers a fixed number of time instants, which are sufficient to understand the dynamics of the system given the largest time gap between causes and effects. This said, it is difficult for an expert to provide or validate a window causal graph because it is difficult to determine which exact time instant is the cause of another. In contrast, there exists another type of causal graph where a node corresponds to a time series, as illustrated in Figure~\ref{fig:full_vs_summary}c, which usually can be analyzed or validated by an expert easily \cite{Wang_2018,Wang_2021,Zhang_2021}. Such graphs are referred to as summary causal graphs \citep{Assaad_2022} and are defined as~follows:
\begin{Definition}[Summary causal graph \cite{Assaad_2022}]
	{Let $\mathcal{X}$ be a multivariate discrete-time series and $\mathcal{G} = (V, E)$ the associated \emph{summary causal graph}. The set of vertices in that graph consists of the set of time series $\mathcal{X}^1,\ldots, \mathcal{X}^d$. The edges $E$ of the graph are defined as follows: variables $\mathcal X^p$ and $\mathcal X^q$ are connected if and only if there exist some time $t$ and some time lag $i$ such that $\mathcal X^p_{t-i}$ causes $\mathcal X^q_{t}$ at time $t$ with a time lag of $0\leq i$ for $p \ne q$ and with a time lag of $0<i$ for $p=q$. 
	}
\end{Definition}
Note that a summary causal graph can be deduced from a window causal graph, but the reverse is not true. 
We focus in this study on the summary causal graph, as it provides a simple and efficient view on the causal relations that exist between time series. In particular, we are interested in inferring a summary causal graph without passing by a window causal graph %please ensure that the original meaning is retained
to avoid unnecessary computations. 
%	Figure~\ref{fig:full_vs_summary} illustrates the difference between a window causal graph (a) and a summary graph (b).

\begin{figure}[H]
	\centering
	\begin{adjustwidth}{-\linewidth}{0cm}
		{\captionsetup{position=bottom,justification=centering}
			\begin{subfigure}{.5\textwidth}
				\centering 
				\begin{tikzpicture}[{black, circle, draw, inner sep=0}]
				\tikzset{nodes={draw,rounded corners},minimum height=0.8cm,minimum width=0.8cm, font=\footnotesize}	
				
				\node (X12) at (0,-0.2) {$X^{1}_{t-2}$} ;
				\node (X11) at (2,-0.2) {$X^{1}_{t-1}$};
				\node (X1) at (4,-0.2) {$X^{1}_{t}$};
				\node (X10) at (6,-0.2) {$X^{1}_{t+1}$};
				\node (X22) at (0,1) {$X^{2}_{t-2}$} ;
				\node(X21) at (2,1) {$X^{2}_{t-1}$};
				\node (X2) at (4,1) {$X^{2}_{t}$};
				\node (X20) at (6,1) {$X^{2}_{t+1}$};
				
				\draw[->,>=latex] (X12) -- (X21);
				\draw[->,>=latex] (X12) -- (X2);
				
				\draw[->,>=latex] (X12) -- (X11);
				\draw[->,>=latex] (X11) -- (X1);
				\draw[->,>=latex] (X1) -- (X10);
				
				\draw[->,>=latex] (X22) -- (X21);
				\draw[->,>=latex] (X21) -- (X2);
				\draw[->,>=latex] (X2) -- (X20);
				
				\draw[->,>=latex] (X11) -- (X2);
				\draw[->,>=latex] (X11) -- (X20);
				\draw[->,>=latex] (X1) -- (X20);
				
				\coordinate[left of=X12] (d1);
				\draw [dashed,>=latex] (X12) to[right] (d1);
				\coordinate[left of=X22] (d1);
				\draw [dashed,>=latex] (X22) to[right] (d1);
				\coordinate[right of=X10] (d1);
				\draw [dashed,>=latex] (X10) to[right] (d1);
				\coordinate[right of=X20] (d1);
				\draw [dashed,>=latex] (X20) to[right] (d1);
				
				\end{tikzpicture}
				\caption{}
				\label{fig:full_vs_summary_a}
			\end{subfigure}
			\hfill
			\begin{subfigure}{.25\textwidth}
				\centering 
				\begin{tikzpicture}[{black, circle, draw, inner sep=0}]
				\tikzset{nodes={draw,rounded corners},minimum height=0.8cm,minimum width=0.8cm, font=\footnotesize}	
				
				\node (X12) at (0,-0.2) {$X^{1}_{t-2}$} ;
				\node (X11) at (2,-0.2) {$X^{1}_{t-1}$};
				\node (X1) at (4,-0.2) {$X^{1}_{t}$};
				\node (X22) at (0,1) {$X^{2}_{t-2}$} ;
				\node(X21) at (2,1) {$X^{2}_{t-1}$};
				\node (X2) at (4,1) {$X^{2}_{t}$};
				
				\draw[->,>=latex] (X12) -- (X21);
				\draw[->,>=latex] (X12) -- (X2);
				
				\draw[->,>=latex] (X12) -- (X11);
				\draw[->,>=latex] (X11) -- (X1);
				
				\draw[->,>=latex] (X22) -- (X21);
				\draw[->,>=latex] (X21) -- (X2);
				
				\draw[->,>=latex] (X11) -- (X2);
				
				\end{tikzpicture}
				\caption{}
				\label{fig:full_vs_summary_b}
			\end{subfigure}
			\hfill
			\begin{subfigure}{.25\textwidth}
				\centering 
				\begin{tikzpicture}[{black, circle, draw, inner sep=0}]
				\tikzset{nodes={draw,rounded corners},minimum height=0.8cm,minimum width=0.8cm, font=\footnotesize}	
				
				\node (X1) at (2,0) {${X}^{1}$} ;
				\node (X3) at (0,0) {${X}^{2}$};
				%		\node (X2) at (0,1) {${X}^{2}$};
				
				\draw[->,>=latex] (X3) -- (X1);
				%		\draw[->,>=latex] (X3) -- (X2);
				
				\draw[->,>=latex] (X1) to [out=0,in=45, looseness=2] (X1);
				%		\draw[->,>=latex] (X2) to [out=0,in=45, looseness=2] (X2);
				\draw[->,>=latex] (X3) to [out=180,in=135, looseness=2] (X3);
				\end{tikzpicture}
				\caption{}
				\label{fig:full_vs_summary_c}
		\end{subfigure}}
	\end{adjustwidth}
	\caption{Example of a full-time causal graph (\textbf{a}), a window causal graph (\textbf{b}), and a summary causal graph (\textbf{c}).}\label{fig:full_vs_summary}
\end{figure}

An important aspect of real-world time series is that different time series, as they measure different elements, usually have different sampling rates. Despite this, the algorithms that have been developed so far to discover causal structures \citep{Granger_1969, Peters_2013, Runge_2019, Nauta_2019} rely on the idealized assumptions that all time series have the same sampling rates with identical timestamps ({assuming identical timestamps in the case of identical sampling rates seems reasonable as one can shift time series so that they coincide in time}). %MDPI: The footnote is not allowed, moved into the text, please conform. (Confirmed)

We introduce in this paper a constraint-based strategy to infer a summary causal graph from discrete-time series with continuous values with equal or different sampling rates under the two classical assumptions of causal discovery (causal Markov condition, faithfulness), in addition to acyclicity in summary causal graphs.
In summary, our contribution is four-fold:
\begin{itemize}
	%			\item First of all, we show that directly addressing the problem of learning summary causal graphs, without resorting to window causal graphs, can be beneficial;
	\item First of all, we propose a new causal temporal mutual information measure defined on a window-based representation of time series;
	\item We then show how this measure relates to an entropy reduction principle, which can be seen as a special case of the probability raising principle;
	\item We also show how this measure can be used for time series with different sampling~rates;
	\item We finally combine these three ingredients in PC-like and FCI-like algorithms \citep{Spirtes_2000} to construct the summary causal graph from time series with equal or different sampling~rates.
\end{itemize}

The remainder of the paper is organized as follows: Section \ref{sec:rw} describes the related work. Section \ref{sec:chapEntropyBased_mu} introduces the (conditional) mutual information measures we propose for time series and the entropy reduction principle that our method is based on. Section~\ref{sec:chapEntropyBased_cd} presents two causal discovery algorithms we developed on top of these measures.
The causal discovery algorithms we propose are illustrated and evaluated on datasets, including time series with equal and different sampling rates and a real dataset in Section~\ref{sec:chapEntropyBased_exp}.
Finally, Section~\ref{sec:chapEntropyBased_concl} concludes the paper.
%%%%%%%%%%%%%%%%%%%%%%%%%%%%%%%%%%%%%%%%%%
\section{Related Work} \label{sec:rw}
Granger Causality is one of the oldest methods proposed to detect causal relations between time series. However, in its standard form \citep{Granger_1969}, it is known to handle a restricted version of causality that focuses on linear relations and causal priorities, as it assumes that the past of a cause is necessary and sufficient for optimally forecasting its effect. This approach has nevertheless been improved since then in \texttt{MVGC} \citep{Granger_2004} and has recently been extended to handle nonlinearities through an attention mechanism within convolutional networks \citep{Nauta_2019}. This last extension is referred to as \texttt{TCDF}.
{Score-based approaches \cite{Chickering_2002} search over the space of possible graphs, trying to maximize a score that reflects how well the graph fits the data. Recently, a new score-based method called \texttt{Dynotears} \cite{Pamfil_2020} was presented to infer a window causal graph from time series.}
In a different line, approaches based on the noise assume that the causal system can be defined by a set of equations that explains each variable by its direct causes and an additional noise. Causal relations are in this case discovered using footprints produced by the causal asymmetry in the data. For time series, the most popular nonlinear algorithm in this family is \texttt{TiMINo} \citep{Peters_2013}, which discovers a causal relationship by looking at independence between the noise and the potential causes. %The main drawbacks of these approaches are the need of a large sample size to achieve good performance and the simplifying assumptions they make on the relations between causes and effects \citep{Malinsky_2018_2}.
The most popular approaches for inferring causal graphs are certainly constraint-based approaches, based on the PC and FCI algorithms \cite{Spirtes_2000}, which turn knowledge about (conditional) independencies into causal knowledge assuming the causal Markov condition and faithfulness. Several algorithms, adapted from non-temporal causal graph discovery algorithms, have been proposed in this family for time series, namely \texttt{oCSE} by \cite{Sun_2015}, which is limited to one time lag between causal relations, \texttt{PCMCI} \cite{Runge_2019}, and \texttt{tsFCI} \cite{Entner_2010}. Our work is thus more closely related to that of \cite{Runge_2019,Entner_2010}, which use constraint-based strategies to infer a window causal graph. However, we focus here on the summary causal graph, and we introduce a new causal temporal mutual information measure and entropy reduction principle on which we ground our causal discovery algorithms.
Constraint-based methods have been also used jointly with other approaches (e.g., {with a score-based method \cite{Malinsky_2018}} and with a noise-based method \citep{Assaad_2021}), but we consider such hybrid methods as beyond the scope of this paper.
%It uses a noise based strategy to detect the causal order between the time series that can represent a complete oriented graph, then uses a constraint based strategy to prune unnecessary edges.

At the core of constraint-based approaches lie (in)dependence measures, to detect relevant dependencies, which are based here on an information theoretic approach. 
Since their introduction \citep{Shannon_1948}, information theoretic measures have become very popular due to their non-parametric nature, their robustness against strictly monotonic transformations, which makes them capable of handling nonlinear distortions in the system, and their good behavior in previous studies on causal discovery \citep{Affeldt_2015}.
%to their non parametric nature and their robustness. 
However, their application to temporal data raises several problems related to the fact that time series may have different sampling rates, be shifted in time, and have strong internal dependencies. Many studies have attempted to re-formalize mutual information for time series. {Reference} \cite{Galka_2006} considered each value of each time series as different random variables and proceeded by whitening the data, such that time-dependent data will be transformed into independent residuals through a parametric model. However, whitening the data can have severe consequences on causal relations. {Reference} %MDPI: newly added information, please confirm. (Confirmed)
\cite{Schreiber_2000} proposed a reformulation of mutual information, called the transfer entropy, which represents the information flow from one state to another and, thus, is asymmetric. Later, {\cite{Sun_2015} generalized transfer entropy to handle conditioning. However, their approach, called causation entropy, can only handle causal relations with lags equal to $1$. Closely related, the directed information \cite{Marko_1973, Massey_1990} allows the computation of the mutual information between two instantaneous time series conditioned on the past of a third time series with a lag of $1$. One of our goals in this paper is to extend these approaches to handle any lag equal to or greater than $0$.}
%Inspired by \cite{Kraskov_2004}, reference \cite{Frenzel_2007} proposed a formulation where time series are represented by vectors and estimated the mutual information assuming that all vectors are statistically independent. This said, time series are still assumed to have equal sampling rates. 
Another related approach is the time-delayed mutual information proposed in~\cite{Albers}, which aims at addressing the problem of non-uniform sampling rates. The computation of the time-delayed mutual information relates single points from a single time series (shifted in time), but does not consider potentially complex relations between time stamps in different time series, as we do through the use of window-based representations and compatible time lags. %The time-delayed mutual information can be seen as a special case of the causal temporal mutual information we introduce in the next section, by considering windows of size 1 and a single time series.
The measure we propose is more suited to discovering summary causal graphs as it can consider potentially complex relations between timestamps in different time series through the use of window-based representations and compatible time lags and is more general as it can consider different sampling rates.

%Since their introduction \citep{Shannon_1948}, information theoretic measures have become very popular. 
%However, their application to temporal data raises several problems related to the fact that time series may have different sampling rates, can be shifted in time and may have strong internal dependencies. 
%Many studies have attempted to re-formalize mutual information for time series: \cite{Galka_2006} decorrelated observations by whitening data (which may have severe consequences on causal relations); 
%\cite{Schreiber_2000} represents the information flow from one state to another within the transfer entropy, which is thus asymmetric;
%\cite{Frenzel_2007}, inspired by \cite{Kraskov_2004}, represented time series by vectors that are assumed to be statistically independent; 
%the Time Delayed Mutual Information proposed in \cite{Albers}, closer to our proposal\footnote{It can be seen as a special case of our measure by considering windows of size 1 and a single time series.}, aims at addressing the problem of non uniform sampling rates. 
%The measure we propose is more suited to discover summary graph as it can consider potentially complex relations between timestamps in different time series through the use of window-based representations and compatible time lags, and is more general as it can consider different sampling rate. 

\section{Information Measures for Causal Discovery in Time Series} \label{sec:chapEntropyBased_mu}

We present in this section a new 
%method for causal discovery in time series inspired by the PC algorithm \citep{Spirtes_2000}. It also makes use of the Probability Raising Principle (PRP, \citet{Suppes_1970}) and is based on
mutual information measure that operates on a window-based representation of time series to assess whether time series are (conditionally) dependent or not. We then show how this measure is related to an entropy reduction principle that is a special case of the probability raising principle \citep{Suppes_1970}.

We first assume that all time series are aligned in time, with the same sampling rate, prior to showing how our development can be applied to time series with different sampling rates. Without loss of generality, time instants are assumed to be integers. Lastly, as performed in previous studies \citep{Schreiber_2000}, we assume that all time series are first-order Markov self-causal (any time instant is caused by its previous instant within the same time series).

\subsection{Causal Temporal Mutual Information}

Let us consider {\small{$d$}} univariate time series {\small{$X^{1}, \cdots, X^{d}$}} and their observations $(X_t^p)_{1\le t \le N_p, 1\le p \le d}$. Throughout this section, we will make use of the following Example \ref{ex1}, illustrated in Figure~\ref{fig:full_vs_summary}, to discuss the notions we introduce.

%	Let us consider 	{\small{$d$}} univariate time series 	{\small{$X^{1}, \cdots, X^{d}$}}, and their observations 	{\small{$(v^{p}_{1}, \cdots, v^{p}_{N_p})$ ($1 \le p \le d$)}}, where 	{\small{$v^{p}_{t}$ ($1 \le t \le N_p$)}} is the value for the $p$-th time series at time index 	{\small{$t$}} and {\small{$N_p$}} is the length of {\small{$X^{p}$}}. Throughout this section, we will make use of the following example, illustrated in Figure~\ref{fig:WindowLag}, to discuss the notions we introduce.

\begin{Example}\label{ex1}
	Let us consider the following two time series defined by, for all $t$,
	%
	%\begin{align*}
	%\forall t: \qquad X^{p}_t = \xi^{p}_t, \,\,\,\, X^{q}_t = X^{p}_{t-2}+X^{p}_{t-1} +\xi^{q}_t,
	%\end{align*}
	{\small{
			\begin{align*}
			X^{1}_t &= X^{1}_{t-1} + \xi^{1}_t,\\
			X^{2}_t &= X^{2}_{t-1} + X^{1}_{t-2}+X^{1}_{t-1} +\xi^{2}_t,
			\end{align*}
	}}
	with {\small{$(\xi^{1}_t, \xi^{2}_t) \sim \mathcal{N}(0,1)$}}.
\end{Example}

One can see in Example~\ref{ex1} that, in order to capture the dependencies between the two time series, one needs to take into account a lag between them, as the true causal relations are not instantaneous. Several studies have recognized the importance of taking into account lags to measure (conditional) dependencies between time series; for example, in \cite{Runge_2019}, a pointwise mutual information between time series with lags was used to assess whether they are dependent or not. 

In addition to lags, Example~\ref{ex1} also reveals that a window-based representation may be necessary to fully capture the dependencies between the two time series. Indeed, as $X^{2}_{t-1}$ and $X^{2}_t$ are the effects of the same cause ($X^{1}_{t-2}$), it may be convenient to consider them together when assessing whether the time series are dependent or not. For example, defining (overlapping) windows of size two for $X^{2}$ and one for $X^{1}$ with a lag of 1 from $X^{1}$ to $X^{2}$, as in Figure~\ref{fig:WindowLag}, allows one to fully represent the causal dependencies between the two time series. 

\begin{figure}[H]
	%\centering
	\begin{tikzpicture}[{black, circle, draw, inner sep=0}]
	\tikzset{nodes={draw,rounded corners},minimum height=0.8cm,minimum width=0.8cm, font=\footnotesize}	
	\tikzset{transparent/.append style={opacity=.2}}	
	
	\node (X12) at (0,-0.2) {$X^{1}_{t-2}$} ;
	\node [transparent] (X11) at (2,-0.2) {$X^{1}_{t-1}$};
	\node [transparent] (X1) at (4,-0.2) {$X^{1}_{t}$};
	\node [transparent] (X10) at (6,-0.2) {$X^{1}_{t+1}$};
	\node [transparent] (X22) at (0,1) {$X^{2}_{t-2}$} ;
	\node(X21) at (2,1) {$X^{2}_{t-1}$};
	\node (X2) at (4,1) {$X^{2}_{t}$};
	\node [transparent] (X20) at (6,1) {$X^{2}_{t+1}$};
	
	\draw[->,>=latex] (X12) -- (X21);
	\draw[->,>=latex] (X12) -- (X2);
	
	\draw[->,>=latex, opacity=.2] (X12) -- (X11);
	\draw[->,>=latex, opacity=.2] (X11) -- (X1);
	\draw[->,>=latex, opacity=.2] (X1) -- (X10);
	
	\draw[->,>=latex, opacity=.2] (X22) -- (X21);
	\draw[->,>=latex, opacity=.2] (X21) -- (X2);
	\draw[->,>=latex, opacity=.2] (X2) -- (X20);
	
	\draw[->,>=latex, opacity=.2] (X11) -- (X2);
	\draw[->,>=latex, opacity=.2] (X11) -- (X20);
	\draw[->,>=latex, opacity=.2] (X1) -- (X20);
	
	\draw (1.5, 0.6) rectangle (4.5 ,1.6);
	\draw[opacity= .2] (-0.5, 0.5) rectangle (2.5 ,1.5);
	\draw[opacity= .2] (3.5, 0.5) rectangle (6.5 ,1.5);
	\draw (-0.5, -0.7) rectangle (0.5 ,0.2);
	\draw[opacity= .2] (1.5, -0.7) rectangle (2.5 ,0.3);
	\draw[opacity= .2] (3.5, -0.7) rectangle (4.5 ,0.3);
	\draw[opacity= .2] (5.5, -0.7) rectangle (6.5 ,0.3);
	%\draw[opacity= .2] (5.5, -0.7) rectangle (8.5 ,0.3);
	\end{tikzpicture}
	
	\caption{Why do we need windows and lags? An illustration with two time series where $X^{1}$ causes $X^{2}$ in two steps (circles correspond to observed points and rectangles to windows). The arrows in black are discussed in the text.}\label{fig:WindowLag}
\end{figure}

%	\begin{Definition} \label{def:chapEntropyBased_complag} Let 	{\small{$\gamma_{\max}$}} denote the maximum lag between two time series 	{\small{$X^{p}$}} and 	{\small{$X^{q}$}} and let 	{\small{$\lambda_{\max}=2\gamma_{\max}+1$}}\footnote{\textcolor{red}{
%		Note the factor $2$ in $\lambda_{max}$ is needed to capture all possible causal relations between $X^p$ and $X^q$.} }. 
%		%the largest possible window size. 
%		The window-based representation, of size 	{\small{$0 < \lambda_{pq} \le \lambda_{\max} < N_p$}}, of the time series 	{\small{$X^{p}$}} with respect to 	{\small{$X^{q}$}}, which will be denoted 	{\small{$X^{(p; \lambda_{pq})}$}}, simply amounts to considering 	{\small{$(N_p-\lambda_{pq}+1)$}} windows: 	{\small{$w_t^{(p;\lambda_{pq})}=(v_{t}^{p}, \cdots, v_{t+\lambda_{pq}-1}^{p}), \,\, 1 \le t \le N_p-\lambda_{pq}+1$}}. The window-based representation, of size 	{\small{$0 < \lambda_{qp} \le \lambda_{\max} < N_q$}}, of the time series 	{\small{$X^{q}$}} with respect to 	{\small{$X^{p}$}} is defined in the same way. A temporal lag 	{\small{$\gamma_{pq}\in \mathbb{Z}$}} \emph{compatible} with 	{\small{$\lambda_{pq}$}} and 	{\small{$\lambda_{qp}$}} relates windows in 	{\small{$X^{(p; \lambda_{pq})}$}} and 	{\small{$X^{(q; \lambda_{qp})}$}} in such a way that the starting time of the related windows are separated by 	{\small{$\gamma_{pq}$}}. We denote by 	{\small{$\mathcal{C}^{(p,q)}$}} the set of window sizes and compatible temporal lags.
%	\end{Definition}
\begin{Definition} \label{def:chapEntropyBased_complag} Let {\small{$\gamma_{\max}$}} denote the maximum lag between two time series {\small{$X^{p}$}} and 	{\small{$X^{q}$}}, and let the maximum window size {\small{$\lambda_{\max}=\gamma_{\max}+1$}}. 
	%the largest possible window size. 
	The window-based representation, of size {\small{$0 < \lambda_{pq} \le \lambda_{\max} < N_p$}}, of the time series {\small{$X^{p}$}} with respect to 	{\small{$X^{q}$}}, which will be denoted {\small{$X^{(p; \lambda_{pq})}$}}, simply amounts to considering {\small{$(N_p-\lambda_{pq}+1)$}} windows: 	{\small{$(X_{t}^{p}, \cdots, X_{t+\lambda_{pq}-1}^{p}), \,\, 1 \le t \le N_p-\lambda_{pq}+1$}}. The window-based representation, of size {\small{$0 < \lambda_{qp} \le \lambda_{\max} < N_q$}}, of the time series {\small{$X^{q}$}} with respect to {\small{$X^{p}$}} is defined in the same way. A temporal lag 	{\small{$\gamma_{pq}\in \mathbb{Z}$}} \emph{compatible} with {\small{$\lambda_{pq}$}} and {\small{$\lambda_{qp}$}} relates windows in {\small{$X^{(p; \lambda_{pq})}$}} and {\small{$X^{(q; \lambda_{qp})}$}} with starting time points separated by {\small{$\gamma_{pq}$}}. We denote by {\small{$\mathcal{C}^{(p,q)}$}} the set of window sizes and compatible temporal lags.
\end{Definition}

Based on the above elements, we define the \textit{causal temporal mutual information} between two time series $X^p$ and $X^q$ as the maximum of the standard mutual information over all possible compatible temporal lags and windows {\small{$\mathcal{C}^{(p,q)}$}}, conditioned by the past of the two time series. Indeed, as we are interested in obtaining a summary causal graph, we do not have to consider all the potential dependencies between two time series (which would be necessary for inferring a window causal graph). Using the maximum over all possible associations is a way to summarize all temporal dependencies, which ensures that one does not miss a dependency between the two time series. 
%As mutual information measures the amount of information brought by the observations of one variable on the observations of another variable, taking the maximum 
Furthermore, conditioning on the past allows one to eliminate spurious dependencies in the form of auto-correlation, as in transfer entropy \citep{Schreiber_2000}. We follow this idea here and, as in transfer entropy, consider windows of size $1$ and a temporal lag of $1$ for conditioning on the past, which is in line with the first-order Markov self-causal assumption mentioned above.

%\textcolor{red}{ Condition on all past}

\begin{Definition}%[CTMI]
	Consider two time series $X^{p}$ and $X^{q}$. We define the \emph{causal temporal mutual information} between $X^{p}$ and $X^{q}$ as:
	{\small{
			\begin{align}
			\text{CTMI}(X^{p}; X^{q}) \label{eq:CTMI} = &
			\max_{(\lambda_{pq}, \lambda_{qp}, \gamma_{pq}) \in \mathcal{C}^{(p,q)}} I(X_t^{(p;\lambda_{pq})};X_{t +\gamma_{pq}}^{(q;\lambda_{qp})} |X^{(p;1)}_{t-1},X_{t +\gamma_{pq}-1}^{(q;1)} )\\
			\delequal & I(X_t^{(p;\bar \lambda_{pq})};X_{t +\bar\gamma_{pq}}^{(q;\bar\lambda_{qp})}|X^{(p;1)}_{t -1};X_{t +\bar\gamma_{pq}-1}^{(q;1)}),\nonumber
			\end{align}
	}}
	where $I$ represents the mutual information. In case the maximum can be obtained with different values in $\mathcal{C}^{(p,q)}$, we first set $\bar\gamma_{pq}$ to its largest possible value. We then set $\bar \lambda_{pq}$ to its smallest possible value and, finally, $\bar \lambda_{qp}$ to its smallest possible value.
	%$$\bar\gamma_{pq}= \min_{|\gamma_{pq}|} arg\max_{(\lambda_{pq}, \lambda_{qp}, \gamma_{pq}) \in \mathcal{C}^{(p,q)}} I(X_t^{(p;\lambda_{pq})};X_{t +\gamma_{pq}}^{(q;\lambda_{qp})} |X^{(p;1)}_{t-1},X_{t +\gamma_{pq}-1}^{(q;1)} ),$$ 
	$\bar\gamma_{pq}$,
	$\bar\lambda_{pq}$, and $\bar\lambda_{qp}$, respectively, correspond to the optimal lag and optimal windows. 
	%\textcolor{red}{If there is many $(\bar\lambda_{pq}, \bar\lambda_{qp}, \bar\gamma_{pq})$ that correspond to the maximal mutual information, we choose the smallest $\bar\lambda_{pq}, \bar\lambda_{qp}$ that correspond to the smallest $|\bar\gamma_{pq}|$.}
\end{Definition}

In the context we have retained, in which dependencies are constant over time, CTMI satisfies the standard properties of mutual information, namely it is nonnegative, symmetric, and equals 0 \textit{iff} time series are independent. Thus, two time series $X^{p}$ and $X^{q}$ such that $CTMI(X^{p};X^{q}) > 0$ are dependent. Setting $\bar\gamma_{pq}$ to its largest possible value allows one to get rid of instants that are not crucial in determining the mutual information between two time series. The choice for the window sizes, even though arbitrary on the choice of treating one window size before the other, is based on the same ground, as the mutual information defined above cannot decrease when one increases the size of the windows. Indeed: 
{\small{
		\begin{align}\label{ineq:ctmi}
		& I(X_t^{(p;\lambda_{pq})}; X_{t+\gamma_{pq}}^{(q;\lambda_{qp})} \mid X^{(p;1)}_{t-1}, X^{(q;1)}_{t+\gamma_{pq}-1}) \nonumber \\
		& = I((X_t^{(p;\lambda_{pq}-1)}, X_{t+\lambda_{pq}-1}^{(p;1)}); X_{t+\gamma_{pq}}^{(q;\lambda_{qp})} \mid X^{(p;1)}_{t-1}, X^{(q;1)}_{t+\gamma_{pq}-1}) \nonumber \\
		& = I(X_t^{(p;\lambda_{pq}-1)}; X_{t+\gamma_{pq}}^{(q;\lambda_{qp})} \mid X^{(p;1)}_{t-1}, X^{(q;1)}_{t+\gamma_{pq}-1})
		+ I(X_{t+\lambda_{pq}-1}^{(p;1)}; X_{t+\gamma_{pq}}^{(q;\lambda_{qp})} \mid X^{(p;1)}_{t-1}, X^{(q;1)}_{t+\gamma_{pq}-1}, X_t^{(p;\lambda_{pq}-1)}) \nonumber \\
		& \ge I(X_t^{(p;\lambda_{pq}-1)}; X_{t+\gamma_{pq}}^{(q;\lambda_{qp})} \mid X^{(p;1)}_{t-1}, X^{(q;1)}_{t+\gamma_{pq}-1}).
		\end{align}
}}

The last inequality is due to the fact that mutual information is positive.

It is also interesting to note that, given two time series $X^p$ and $X^q$ such that $X^p$ causes $X^q$ with $\gamma_{pq}=0$, CTMI does not necessarily increase symmetrically with respect to the increase of $\lambda_{pq}$ and $\lambda_{qp}$. For an illustration on a simple model, see Figure~\ref{fig:ctmi_asymetric}. In Figure~\ref{fig:ctmi_asymetric}a, the mutual information conditioned on the past between $X^p$ and $X^q$ with $\gamma_{pq}=0$ is positive since $X^p$ causes $X^q$ instantaneously. In Figure~\ref{fig:ctmi_asymetric}b the same mutual information does not increase when increasing only the window size of the cause. 
However, the mutual information increases when increasing only the window size of the effect, as in Figure~\ref{fig:ctmi_asymetric}c, or when increasing simultaneously the window sizes of the effect and the cause, as in Figure~\ref{fig:ctmi_asymetric}d. 
This property can be used to early stop the increase of the window size. For instance, if for a fixed lag, increasing the window size of one time series increases the (conditional) mutual information between them but increasing the window size of the other time series does not increase the (conditional) mutual information then one should stop investigating bigger window sizes. In addition, in some situations, as will be shown in Section~\ref{sec:PC}, when given two-time series that are directly causally related (without knowing which one is the cause and which one is the effect) this property can be used to distinguish which one is the cause and which one is the effect.

\begin{figure}[H]
	%\centering
	{\captionsetup{position=bottom,justification=centering}\hspace{-18pt}\begin{subfigure}{0.45\linewidth}
			\centering
			\vspace{-1cm}
			\begin{tikzpicture}[{black, circle, draw, inner sep=0}]
			\tikzset{nodes={draw,rounded corners},minimum height=0.8cm,minimum width=0.8cm, font=\footnotesize}	
			\tikzset{transparent/.append style={opacity=.2}}	
			
			\node (X10) at (0,-0.5) {$X^{p}_{t-1}$} ;
			\node (X11) at (2,-0.5) {$X^{p}_{t}$};
			\node[transparent] (X12) at (4,-0.5) {$X^{p}_{t+1}$};
			\node (X20) at (0,1) {$X^{q}_{t-1}$} ;
			\node (X21) at (2,1) {$X^{q}_{t}$};
			\node[transparent] (X22) at (4,1) {$X^{q}_{t+1}$};
			\draw[->,>=latex] (X10) -- (X11);
			\draw[->,>=latex, opacity=.2] (X11) -- (X12);
			\draw[->,>=latex] (X20) -- (X21);
			\draw[->,>=latex, opacity=.2] (X21) -- (X22);
			
			\draw[->,>=latex,opacity=.2] (X10) -- (X20);
			\draw[->,>=latex,line width=0.05cm] (X11) -- (X21);
			\draw[->,>=latex,opacity=.2] (X12) -- (X22);
			
			\draw (1.5, -1.0) rectangle (2.5 ,-0.0);
			\draw (1.5, 0.5) rectangle (2.5 ,1.5);
			\end{tikzpicture}	
			\caption{}
			\label{fig:ctmi_asymetric_a}
			\vspace{0.9cm}
		\end{subfigure}
		\hfill 	
		\begin{subfigure}[t]{0.45\linewidth}
			\centering 
			\begin{tikzpicture}[{black, circle, draw, inner sep=0}]
			\tikzset{nodes={draw,rounded corners},minimum height=0.8cm,minimum width=0.8cm, font=\footnotesize}	
			\tikzset{transparent/.append style={opacity=.2}}	
			
			\node (X10) at (0,-0.5) {$X^{p}_{t-1}$} ;
			\node (X11) at (2,-0.5) {$X^{p}_{t}$};
			\node (X12) at (4,-0.5) {$X^{p}_{t+1}$};
			\node (X20) at (0,1) {$X^{q}_{t-1}$} ;
			\node (X21) at (2,1) {$X^{q}_{t}$};
			\node[transparent] (X22) at (4,1) {$X^{q}_{t+1}$};
			\draw[->,>=latex] (X10) -- (X11);
			\draw[->,>=latex] (X11) -- (X12);
			\draw[->,>=latex] (X20) -- (X21);
			\draw[->,>=latex, opacity=.2] (X21) -- (X22);
			
			\draw[->,>=latex,opacity=.2] (X10) -- (X20);
			\draw[->,>=latex,line width=0.05cm] (X11) -- (X21);
			\draw[->,>=latex,opacity=.2] (X12) -- (X22);
			
			\draw (1.5, -1.) rectangle (4.5 ,-0.0);
			\draw (1.5, 0.5) rectangle (2.5 ,1.5);
			\end{tikzpicture}	
			\caption{}
			\label{fig:ctmi_asymetric_b}
		\end{subfigure}
		\vspace{0.1cm}
		
		\hspace{-18pt}\begin{subfigure}[t]{0.45\linewidth}
			\centering 
			\begin{tikzpicture}[{black, circle, draw, inner sep=0}]
			\tikzset{nodes={draw,rounded corners},minimum height=0.8cm,minimum width=0.8cm, font=\footnotesize}	
			\tikzset{transparent/.append style={opacity=.2}}	
			
			\node (X10) at (0,-0.5) {$X^{p}_{t-1}$} ;
			\node (X11) at (2,-0.5) {$X^{p}_{t}$};
			\node[transparent] (X12) at (4,-0.5) {$X^{p}_{t+1}$};
			\node (X20) at (0,1) {$X^{q}_{t-1}$} ;
			\node (X21) at (2,1) {$X^{q}_{t}$};
			\node (X22) at (4,1) {$X^{q}_{t+1}$};
			\draw[->,>=latex] (X10) -- (X11);
			\draw[->,>=latex, opacity=.2] (X11) -- (X12);
			\draw[->,>=latex] (X20) -- (X21);
			\draw[->,>=latex] (X21) -- (X22);
			
			\draw[->,>=latex,opacity=.2] (X10) -- (X20);
			\draw[->,>=latex,line width=0.05cm] (X11) -- (X21);
			\draw[->,>=latex,opacity=.2] (X12) -- (X22);
			
			\draw[dashed,line width=0.05cm] (X11) -- (X22);
			
			\draw (1.5, -1.) rectangle (2.5 ,-0.0);
			\draw (1.5, 0.5) rectangle (4.5 ,1.5);
			
			\end{tikzpicture}	
			\caption{}
			\label{fig:ctmi_asymetric_c}
		\end{subfigure}
		\hfill 
		\begin{subfigure}[t]{0.45\linewidth}
			\centering 
			\begin{tikzpicture}[{black, circle, draw, inner sep=0}]
			\tikzset{nodes={draw,rounded corners},minimum height=0.8cm,minimum width=0.8cm, font=\footnotesize}	
			\tikzset{transparent/.append style={opacity=.2}}	
			
			\node (X10) at (0,-0.5) {$X^{p}_{t-1}$} ;
			\node (X11) at (2,-0.5) {$X^{p}_{t}$};
			\node (X12) at (4,-0.5) {$X^{p}_{t+1}$};
			\node (X20) at (0,1) {$X^{q}_{t-1}$} ;
			\node (X21) at (2,1) {$X^{q}_{t}$};
			\node (X22) at (4,1) {$X^{q}_{t+1}$};
			\draw[->,>=latex] (X10) -- (X11);
			\draw[->,>=latex] (X11) -- (X12);
			\draw[->,>=latex] (X20) -- (X21);
			\draw[->,>=latex] (X21) -- (X22);
			
			\draw[->,>=latex,opacity=.2] (X10) -- (X20);
			\draw[->,>=latex,line width=0.05cm] (X11) -- (X21);
			\draw[->,>=latex,line width=0.05cm] (X12) -- (X22);
			
			\draw (1.5, -1.0) rectangle (4.5 ,-0.0);
			\draw (1.5, 0.5) rectangle (4.5 ,1.5);
			
			\end{tikzpicture}	
			\caption{}
			\label{fig:ctmi_asymetric_d}
	\end{subfigure}}
	\caption{{Illustration} of the asymmetric increase of CTMI with the increase of the window sizes. The mutual information conditioned on the past in (\textbf{a}) is ${\displaystyle I(X^p_t, X^q_t\mid X^p_{t-1}, X^q_{t-1}) > 0}$. It increases when increasing only the window size of the effect, as in (\textbf{c}), i.e, $I(X^{p}_t, X^{(q;2)}_t\mid X^p_{t-1}, X^q_{t-1}) > I(X^p_t, X^q_t\mid X^p_{t-1}, X^q_{t-1})$, or when increasing simultaneously the window sizes of the effect and the cause, as in (\textbf{d}), i.e, $ I(X^{(p;2)}_t, X^{(q;2)}_t\mid X^p_{t-1}, X^q_{t-1}) > I(X^{p}_t, X^{(q;2)}_t\mid X^p_{t-1}, X^q_{t-1})$. However, it does not increase when increasing only the window size of the cause, as in (\textbf{b}), i.e, $I(X^{(p;2)}_t, X^{q}_t\mid X^p_{t-1}, X^q_{t-1}) = I(X^p_t, X^q_t\mid X^p_{t-1}, X^q_{t-1})$. Dashed lines are for correlations that are not causations, and bold arrows correspond to causal relations between the window representations of time series.}\label{fig:ctmi_asymetric}%MDPI: moved subfigure's explanation into caption, please confirm. (Confirmed and reformulated)
\end{figure}

\begin{Example}\label{exple:tmi}
	Consider the structure described in Example \ref{ex1}, and assume that $\lambda_{\max}= 3$.
	First, we have, for the standard mutual information, 
	{\small{$$I(X_t^{(1;1)}; X_{t}^{(2;1)} \mid X^{(1;1)}_{t-1}, X^{(2;1)}_{t-1}) = 0.$$}}
	We also have that any $\gamma_{12}<0$ has zero mutual information because conditioning on the past of $X_t^{(1;1)}; X_{t}^{(2;1)}$ (namely $X^{(1;1)}_{t-1}; X_{t-1}^{(2;1)}$) is closing all paths from $X_{t-i}^{(2;1)}$ to $X_t^{(1;1)}$ for all $i>0$. 
	For $\gamma_{12}>0$, starting by $\gamma_{12}=1$, 
	{\small{$$I(X_t^{(1;1)}; X_{t+1}^{(2;1)} \mid X^{(1;1)}_{t-1}, X^{(2;1)}_{t}) > 0.$$}}
	This is similar for $\gamma_{12}>2$. %however any $\gamma>2$ will again have a zero mutual information.
	Now, any increase of $\lambda_{21}$ alone or of $\lambda_{12}$ and $\lambda_{21}$ will generate an increase in the mutual information as long as the difference between the last time point of the window of $X^1$ (the cause) and the last time point of the window of $X^2$ is less than or equal to $\gamma_{12}$ as
	{\small{
			\begin{align*}
			I(X_t^{(1;\lambda_{12}-1)}; X_{t+\gamma_{12}}^{(2;\lambda_{12}-1-\gamma_{12})} \mid X^{(1;1)}_{t-1}, X^{(2;1)}_{t+\gamma_{12}-1}) 
			= I(X_t^{(1;\lambda_{12})}; X_{t+\gamma_{12}}^{(2;\lambda_{12}-1-\gamma_{12})} \mid X^{(1;1)}_{t-1}, X^{(2;1)}_{t+\gamma_{12}-1})
			\end{align*}}}
	because 
	{\small{$$I(X_{t+\lambda_{12} -1}^{(1;1)}; X_{t+\gamma_{12}}^{(2;\lambda_{12}-1-\gamma_{12})} \mid X^{(1;1)}_{t-1}, X^{(2;1)}_{t+\gamma_{12}-1}, X_t^{(1;\lambda_{12}-1)}) = 0,$$}}
	where $\gamma_{12}\ge1$ ($1$ is the minimal lag that generates a correlation that cannot be removed by conditioning on the past of $X^1$ and $X^2$).
	%However, increasing the size of $\lambda_{qp}$ would increase CTMI as
	%$$I(X_t^{(p;\lambda_{pq}-1)}; X_{t+\lambda_{qp}+\gamma_{pq}-1}^{(q;1)} \mid X^{(p;1)}_{t-1}, X^{(q;1)}_{t+\gamma_{pq}-1}, X_{t+\gamma_{pq}}^{(q;\lambda_{qp})}) > 0 $$
	%and by consequence increasing $\lambda_{pq}$ after $\lambda_{qp}$ would also increase the mutual CTMI.
	%
	%As one can notice $\lambda_{qp}$ will be equal to $\lambda_{\max} - \gamma_{pq}$
	For $\lambda_{\max}=3$, the optimal window size $\bar\lambda_{12}$ is equal to $2$ as $X^{1}$ has no other cause than itself; $\bar\lambda_{21}$ is equal to $2$ as $X^{1}$ causes only (except itself) $X^{2}_{t+1}$ and $X^{2}_{t}$. %have one common dependence in \textcolor{red}{while $X^{q}_{t+1}$ and $X^{q}_{t-1}$ have none}.
	Furthermore, $\bar\gamma_{12}=1$ and 
	{\small{
			\begin{align*}
			\text{CTMI}(X^{1}; X^{2}) &= I((X^{1}_{t}, X^{1}_{t+1}); (X^{2}_{t+1}, X^{2}_{t+2}) \mid X^{1}_{t-1}, X^{2}_{t})\\ 
			&= I(X^{1}_{t}; (X^{2}_{t+1}, X^{2}_{t+2}) \mid X^{1}_{t-1}, X^{2}_{t}) 
			+ I(X^{1}_{t+1}; (X^{2}_{t+1}, X^{2}_{t+2}) \mid X^{1}_{t-1}, X^{2}_{t}, X^{1}_{t}) \\
			&= I(X^{1}_{t}; X^{2}_{t+1} \mid X^{1}_{t-1}, X^{2}_{t}) + I(X^{1}_{t}; X^{2}_{t+2} \mid X^{1}_{t-1}, X^{2}_{t}, X^{2}_{t+1})\\
			&\quad+ I(X^{1}_{t+1}; X^{2}_{t+1} \mid X^{1}_{t-1}, X^{2}_{t}, X^{1}_{t}) 
			+ I(X^{1}_{t+1}; X^{2}_{t+2} \mid X^{1}_{t-1}, X^{2}_{t}, X^{1}_{t}, X^{2}_{t+1})\\
			&= 2I(X^{1}_{t}; X^{2}_{t+1} \mid X^{1}_{t-1}, X^{2}_{t}) + I(X^{1}_{t}; X^{2}_{t+2} \mid X^{1}_{t-1}, X^{2}_{t}, X^{2}_{t+1})
			= 3\log(3)/4.
			\end{align*}}}
\end{Example}

\subsection{Entropy Reduction Principle}

Interestingly, CTMI can be related to a version of the \textit{probability raising principle} \cite{Suppes_1970}, which states that a cause, here a time series, raises the probability of any of its effects, here another time series, even when the past of the two time series is taken into account, meaning that the relation between the two time series is not negligible compared to the internal dependencies of the time series. In this context, the following definition generalizes to window-based representations of time series the standard definition of {prima facie} causes for discrete variables. % based on the Probability Raising Principle (PRP), as given in \cite{Suppes_1970}.

%\textcolor{red}{condition on all past}

\begin{Definition}[\textbf{{Prima} facie cause for window-based time series}]\label{def:chapEntropyBased_prima} %MDPI: Please confirm if the bold should be retained. (Confirmed)
	Let $X^{p}$ and $X^{q}$ be two time series with window sizes $\lambda_{pq}$ and $\lambda_{qp}$, and let $P_{t,t'}=(X^{(p;1)}_{t-1},X^{(q;1)}_{t'-1})$ represent the past of $X^{p}$ and $X^{q}$ for any two instants $(t,t')$. We say that $X^{p}$ is a \emph{prima facie cause} of $X^{q}$ with delay $\gamma_{pq} > 0$ iff there exist Borel sets $B_p$, $B_q$, and $B_P$ such that one has:
	{\small{$P(X^{(q;\lambda_{qp})}_{t+\gamma_{pq}} \in B_q|X^{(p;\lambda_{pq})}_t \in B_p,P_{t,t+\gamma_{pq}} \in B_P) > P(X^{(q;\lambda_{qp})}_{t+\gamma_{pq}} \in B_q|P_{t,t+\gamma_{pq}} \in B_P).$}}
	%	\small{
	%		%
	%		\begin{align}
	%		P(X^{(q;\lambda_{qp})}_{t+\gamma_{pq}} \in B_q|X^{(p;\lambda_{pq})}_t \in B_p,P_{t,t+\gamma_{pq}} \in B_P) > \nonumber 
	%		 P(X^{(q;\lambda_{qp})}_{t+\gamma_{pq}} \in B_q|P_{t,t+\gamma_{pq}} \in B_P). \nonumber
	%		\end{align}
	%		%
	%	}
\end{Definition}

We now introduce a slightly different principle based on the causal temporal mutual information, which we refer to as the \textit{entropy reduction principle}. %It takes the form below.
\begin{Definition}[\textbf{{Entropic} prima facie cause}]
	\label{def:entropic_prima_facie_cause}
	Using the same notations as in Definition~\ref{def:chapEntropyBased_prima}, we say that $X^{p}$ is an \emph{entropic prima facie cause} of $X^{q}$ with delay $\gamma_{pq} > 0$ iff $I(X_t^{(p;\lambda_{pq})};X_{t +\gamma_{pq}}^{(q;\lambda_{qp})} |P_{t,t+\gamma_{pq}} ) > 0$.
\end{Definition}
Note that considering that the above mutual information is positive is equivalent to considering that the entropy of $X^{q}$ when conditioned on the past reduces when one further conditions on $X^{p}$. 
%The interesting point with this last definition is that it no longer relies on particular Borel sets. Because of that, entropic prima facie causes are easier to identify than prima facie causes. 
One has the following relation between the entropy reduction and the probability raising principles.%\emph{entropic prima facie} and {prima facie} causes.
\begin{Property}
	\label{prop:entropic_prima_facie_cause}
	With the same notations, if $X^{p}$ is an \emph{entropic prima facie} cause of $X^{q}$ with delay $\gamma_{pq} > 0$, then $X^{p}$ is a {prima facie} cause of $X^{q}$ with delay $\gamma_{pq} > 0$. Furthermore, if $\text{CTMI}(X^{p}; X^{q}) > 0$ with $\bar\gamma_{pq} > 0$, then $X^{p}$ is an \emph{entropic prima facie} cause of $X^{q}$ with delay $\bar\gamma_{pq}$.
\end{Property}
%Proof in Appendix~\ref{appendix:proof}.
%
\begin{proof}
	Let us assume that $X^{p}$ is not a {prima facie} cause of $X^{q}$ for the delay $\gamma_{pq}$. Then, for all Borel sets $B_p$, $B_q$, and $B_P$, one has $P(X^{(q;\lambda_{qp})}_{t+\gamma_{pq}} \in B_q|X^{(p;\lambda_{pq})}_t \in B_p,P_{t,t+\gamma_{pq}} \in B_P) \le P(X^{(q;\lambda_{qp})}_{t+\gamma_{pq}} \in B_q|P_{t,t+\gamma_{pq}} \in B_P)$. This translates, in terms of density functions denoted $f$, as: 
	$$\forall (x^{p}_t, x^{q}_{t+\gamma_{pq}}, p_{t,t+\gamma_{pq}}), \,\, f(x^{q}_{t+\gamma_{pq}} | x^{p}_t, p_{t,t+\gamma_{pq}}) \le f(x^{q}_{t+\gamma_{pq}}|p_{t,t+\gamma_{pq}}),$$ 
	which implies that $H(X^{(q;\lambda_{qp})}_{t+\gamma_{pq}} \in B_q|X^{(p;\lambda_{pq})}_t \in B_p,P_{t,t+\gamma_{pq}} \in B_P)$ is greater than $H(X^{(q;\lambda_{qp})}_{t+\gamma_{pq}} \in B_q|P_{t,t+\gamma_{pq}} \in B_P)$, so that $X^{p}$ is not an \emph{entropic prima facie} cause of $X^{p}$ with delay $\gamma_{pq}$. By contraposition, we conclude the proof of the first statement. The second statement directly derives from the definition of CTMI.
\end{proof}

%Lastly, one can define a direct cause as any prima facie cause for which , which corresponds to a true causal relation in the context of the 

%	\begin{Property}
%		\label{prop:asymetric_increase}
%		If a time series $X^{p}$ causes a time series $X^{q}$ instantaneously, i.e, $\gamma_{pq} = 0$, then $\text{CTMI}(X^{p}; X^{q}) > 0$ increase asymetrically.
%	\end{Property}
%	
%	\begin{proof}
%		Let 
%	\end{proof}

\subsection{Conditional Causal Temporal Mutual Information}

We now extend the causal temporal mutual information by conditioning on a set of variables. In a causal discovery setting, conditioning is used to assess whether two dependent time series can be made independent by conditioning on connected time series, {i.e.}, time series that are dependent with at least one of the two times series under consideration. Figure~\ref{fig:ex_dctmi} illustrates the case where the dependence between $X^{1}$ and $X^{2}$ is due to spurious correlations originating from common causes. Conditioning on these common causes should lead to the conditional independence of the two time series. Of course, the conditional variables cannot succeed simultaneously in time the two time series under consideration. This leads us to the following definition of the conditional causal temporal mutual information.

\begin{figure}[H]
	%\centering
	{\captionsetup{position=bottom,justification=centering}\hspace{-18pt}\begin{subfigure}{0.4\textwidth}
			\centering
			\begin{tikzpicture}[{black, circle, draw, inner sep=0}]
			\tikzset{nodes={draw,rounded corners},minimum height=0.8cm,minimum width=0.8cm, font=\footnotesize}	
			\tikzset{transparent/.append style={opacity=.2}}	
			
			\node (X12) at (0,-0.5) {$X^{1}_{t-2}$} ;
			\node (X11) at (1.5,-0.5) {$X^{1}_{t-1}$};
			\node [transparent] (X1) at (3,-0.5) {$X^{1}_{t}$};
			\node (X32) at (0,0.5) {$X^{3}_{t-2}$} ;
			\node [transparent] (X31) at (1.5,0.5) {$X^{3}_{t-1}$};
			\node [transparent] (X3) at (3,0.5) {$X^{3}_{t}$};
			\node [transparent] (X22) at (0,1.5) {$X^{2}_{t-2}$} ;
			\node [transparent](X21) at (1.5,1.5) {$X^{2}_{t-1}$};
			\node (X2) at (3,1.5) {$X^{2}_{t}$};
			
			\draw[->,>=latex, opacity=.2] (X11) -- (X1);
			\draw[->,>=latex, opacity=.2] (X32) -- (X31);
			\draw[->,>=latex, opacity=.2] (X31) -- (X3);
			\draw[->,>=latex, opacity=.2] (X22) -- (X21);
			
			\draw[->,>=latex,line width=0.05cm] (X12) -- (X11);
			\draw[->,>=latex,line width=0.05cm] (X21) -- (X2);
			
			\draw[->,>=latex,line width=0.05cm] (X32) -- (X11);
			\draw[->,>=latex, opacity=.2] (X31) -- (X1);
			\draw[->,>=latex,line width=0.05cm] (X32) -- (X2);
			\draw[dashed,line width=0.05cm] (X11) -- (X2);
			\draw[dashed,line width=0.05cm, opacity=.2] (X12) -- (X21);
			
			\draw[opacity= .2] (1, 1.1) rectangle (2 ,2.1);
			\draw (1, -1) rectangle (2 ,0);
			
			\draw[opacity= .2] (-0.5, 1) rectangle (0.5 ,2);
			\draw (2.5, 1) rectangle (3.5 ,2);
			\draw[opacity= .2] (-0.5, -1) rectangle (0.5 ,0);
			\draw[opacity= .2] (2.5, -1) rectangle (3.5 ,0);
			
			%%%%%%%%%%%%
			\end{tikzpicture}
			\caption{}
		\end{subfigure}
		\hfill 
		\begin{subfigure}{0.4\textwidth}
			\centering
			\begin{tikzpicture}[{black, circle, draw, inner sep=0}]
			\tikzset{nodes={draw,rounded corners},minimum height=0.8cm,minimum width=0.8cm, font=\footnotesize}	
			\tikzset{transparent/.append style={opacity=.2}}	
			
			\node (X12) at (5,-1) {$X^{1}_{t-2}$} ;
			\node (X11) at (6.5,-1) {$X^{1}_{t-1}$};
			\node [transparent] (X1) at (8,-1) {$X^{1}_{t}$};
			\node (X32) at (5,0) {$X^{4}_{t-2}$} ;
			\node [transparent] (X31) at (6.5,0) {$X^{4}_{t-1}$};
			\node [transparent] (X3) at (8,0) {$X^{4}_{t}$};
			\node (X42) at (5,1) {$X^{3}_{t-2}$} ;
			\node [transparent] (X41) at (6.5,1) {$X^{3}_{t-1}$};
			\node [transparent] (X4) at (8,1) {$X^{3}_{t}$};
			\node (X22) at (5,2) {$X^{2}_{t-2}$} ;
			\node (X21) at (6.5,2) {$X^{2}_{t-1}$};
			\node (X2) at (8,2) {$X^{2}_{t}$};
			
			\draw[->,>=latex, opacity=.2] (X11) -- (X1);
			\draw[->,>=latex, opacity=.2] (X32) -- (X31);
			\draw[->,>=latex, opacity=.2] (X31) -- (X3);
			\draw[->,>=latex, opacity=.2] (X42) -- (X41);
			\draw[->,>=latex, opacity=.2] (X41) -- (X4);
			\draw[->,>=latex, opacity=.2] (X22) -- (X21);
			\draw[->,>=latex] (X21) -- (X2);
			
			\draw[->,>=latex,line width=0.05cm] (X12) -- (X11);
			\draw[->,>=latex,line width=0.05cm] (X22) -- (X21);
			
			\draw[->,>=latex,line width=0.05cm] (X32) -- (X11);
			\draw[->,>=latex, opacity=.2] (X31) -- (X1);
			\draw[->,>=latex,line width=0.05cm] (X32) -- (X2);
			
			\draw[->,>=latex,line width=0.05cm] (X42) -- (X11);
			\draw[->,>=latex, opacity=.2] (X41) -- (X1);
			\draw[->,>=latex, opacity=.2] (X41) -- (X2);
			\draw[->,>=latex,line width=0.05cm] (X42) -- (X21);
			
			\draw[dashed,line width=0.05cm] (X11) -- (X21);
			\draw[dashed,line width=0.05cm] (X11) -- (X2);
			\draw[dashed,line width=0.05cm, opacity=.2] (X1) -- (X2);
			\draw[dashed,line width=0.05cm, opacity=.2] (X12) -- (X22);
			\draw[dashed,line width=0.05cm, opacity=.2] (X12) -- (X21);

			\draw (6, 1.6) rectangle (8.5 ,2.6);
			\draw (6, -1.5) rectangle (7 ,-0.5);
			
			\draw[opacity= .2] (4.5, 1.5) rectangle (7,2.5);
			\draw[opacity= .2] (4.5, -1.5) rectangle (5.5 ,-0.5);
			\draw[opacity= .2] (7.5, -1.5) rectangle (8.5 ,-0.5);
			
			\end{tikzpicture}
			\caption{}
	\end{subfigure}}
	\caption{{Example} of conditional independence between dependent time series. In (\textbf{a}) the conditioning set contains one time series $X^3$ in addition to the past of $X^1$ and $X^2$. In (\textbf{b}) the conditioning set contains one time series $X^3$ and $X^4$ in addition to the past of $X^1$ and $X^2$. Dashed lines are for correlations that are not causations, and bold arrows correspond to conditioning variables.}%MDPI: please add explanation for subfigure ab. (Done)
	\label{fig:ex_dctmi}
\end{figure}

%\textcolor{red}{condition on all past}

\begin{Definition}\label{def:chapEntropyBased_condCTMI}
	The \emph{conditional causal temporal mutual information} between two time series $X^{p}$ and $X^{q}$ such that $\bar\gamma_{pq}\ge 0$, conditioned on a {set} $X^{\textbf{R}}=\{X^{r_1}, \cdots, X^{r_K}\}$, is given by: %MDPI: please confirm if the bold for R is necessary, and confirm this type in the full text. (Confirmed)
	{\small{
			\begin{align}
			\label{eq:condCTMI}
			\text{CTMI}(X^{p};X^{q}\mid X^{\textbf{R}}) 
			= I(X_t^{(p;\bar\lambda_{pq})};X_{t +\bar\gamma_{pq}}^{(q;\bar\lambda_{qp})} |(X^{(r_k;\bar\lambda_k)}_{t - \bar\Gamma_{k}})_{1\leq k \leq K}, X^{(p;1)}_{t -1},X_{t +\bar\gamma_{pq}-1}^{(q;1)}).
			\end{align}
	}}
	In case the minimum can be obtained with different values, we first set $\bar\Gamma_{k}$ to its largest possible value. We then set $\bar \lambda_{k}$ to its smallest possible value.
	$(\bar\Gamma_{1}, \ldots, \bar\Gamma_{K})$ and $(\bar\lambda_1, \cdots, \bar\lambda_K)$ correspond to the optimal conditional lags and window sizes, which minimize, for $\Gamma_{1}, \ldots, \Gamma_{K} \ge -\bar\gamma_{pq}$:%$\Gamma_{1}, \ldots, \Gamma_{K} \ge \max(0,\mbox{sgn}(\bar\gamma_{pq})|\bar\gamma_{pq}+1|)$: %\textcolor{red}{$\Gamma_{1}, \ldots, \Gamma_{K} \ge -\max(0,\bar\gamma_{pq})$ }
	{\small{
			\begin{align*}
			I \left(X_t^{(p;\bar\lambda_{pq})};X_{t +\bar\gamma_{pq}}^{(q;\bar\lambda_{qp})} | (X^{(r_k;\lambda_k)}_{t - \Gamma_{k}})_{1\leq k \leq K}, X^{(p;1)}_{t -1},X_{t + \bar\gamma_{pq}-1}^{(q;1)} \right).
			\end{align*}
	}}
	%
	%In the remainder, we will denote $(X^{(r_k;1)}_{t - \bar\Gamma_{k}})_{1 \leq k \leq K}$ by $X^{(\textbf{R};1)}_{t - \bar\Gamma}$.
\end{Definition}

By considering the minimum over compatible lags and window sizes, one guarantees that, if there exist conditioning variables that make the two time series independent, they will be found. 
Note that the case in which $\bar\gamma_{qp}<0$ corresponds to $\text{CTMI}(X^{p};X^{q}\mid X^{\textbf{R}})$, {where} $\bar\gamma_{pq}>0$. %MDPI: for sup R, please use same format, italic or not italic. (Condirmed to not italic)
%Note that the same time series with different temporal lags can appear in $X^{(\textbf{R})}$ so that all potential conditioning variables are considered. %Furthermore, the above measure can readily be extended to a single time series $X^{p}$ to detect temporal dependencies within the time series.}%, with $\gamma_{pp}>0$.

Figure \ref{fig:ex_dctmi} illustrates the above on two different examples. On the left, $X^{1}_{t-1}$ is correlated to $X^{2}_t$ as $X^{3}_{t-2}$ is a common cause with a lag of $1$ for $X^{1}$ and a lag of $2$ for $X^{2}$. Conditioning on $X^{3}_{t-2}$ removes the dependency between $X^{1}$ and $X^{2}$. Note that all time series have here a window of size $1$. On the right, $X^{3}$ and $X^{4}$ are common causes of $X^{1}$ and $X^{2}$: $X^{3}$ causes $X^{1}$ and $X^{2}$ with a temporal lag of $1$, which renders $X^{1}$ and $X^{2}$ correlated at the same time point, while $X^{4}$ causes $X^{1}$ and $X^{2}$ with a temporal lag of $1$ and $2$, respectively, which renders $X^{1}$ and $X^{2}$ correlated at lagged time points. The overall correlation between $X^{1}$ and $X^{2}$ is captured by considering a window size of $2$ in $X^{2}$. All other time series have a window size of $1$. By conditioning on both $X^{3}$ and $X^{4}$, $X^{p}$ and $X^{q}$ become independent, assuming we also condition on the past of $X^1$ and $X^2$ to remove the autocorrelation.

\subsection{Estimation and Testing}\label{subsec:estimation_test}
In practice, the success of the CTMI approach (and in fact, any entropy-based approaches) depends crucially on the reliable estimation of the relevant entropies in question from data. This leads to two practical challenges. 
The first one is based on the fact that entropies must be estimated from finite-time series data. 
The second is that to detect independence, we need a statistical test to check if the temporal causation entropy is equal to zero. 

Here, we rely on the $k$-nearest neighbor method \citep{Kraskov_2004, Frenzel_2007} for the estimation of CTMI. 
The distance between two windows considered here is the supremum distance, i.e., the maximum of the absolute difference between any two values in the two windows.
%
%\begin{adjustwidth}{-\linewidth}{0cm}
	%\centering %% If there is a figure in wide page, please release command \centering
	\begin{align*}
	d((X^{(p;\bar\lambda_{pq})}_t, X^{(q;\bar\lambda_{pq})}_{t+\bar\gamma_{pq}})_i, (X^{(p;\bar\lambda_{pq})}_t, X^{(q;\bar\lambda_{pq})}_{t+\bar\gamma_{pq}})_j) 
	=\max_{\substack{0 \le \ell < \lambda_p, \, 0 \le \ell' < \lambda_q}} (&|(X^{(p;\bar\lambda_{pq})}_t)_{i+\ell}
	- (X^{(p;\bar\lambda_{pq})}_t)_{j+\ell'}|, \\
	\qquad\qquad\qquad\qquad&|(X^{(q;\bar\lambda_{qp})}_t)_{i+\ell'}
	- (X^{(q;\bar\lambda_{qp})}_t)_{j+\ell'}|). 
	\end{align*}
%\end{adjustwidth}
%
In the case of the causal temporal mutual information,
we denote by $ \epsilon_{ik}/2$ the distance from 
$$(X^{(p;\bar\lambda_{pq})}_t, X^{(q;\bar\lambda_{pq})}_{t+\bar\gamma_{pq}}, X^{p}_{t-1}, X^{q}_{t+\bar\gamma_{pq}-1})$$
to its $k$-th neighbor, 
$n_{i}^{1, 3}$, $n_{i}^{2,3}$, and $n_{i}^3$ being the numbers of points with a distance strictly smaller than $\epsilon_{ik}/2$ in the subspace 
$$(X^{(p;\bar\lambda_{pq})}_t, X^{p}_{t-1}, X^{q}_{t+\bar\gamma_{pq}-1}),
(X^{(q;\bar\lambda_{pq})}_{t+\bar\gamma_{pq}}, X^{p}_{t-1}, X^{q}_{t+\bar\gamma_{pq}-1})\text{, and }
(X^{p}_{t-1}, X^{q}_{t+\bar\gamma_{pq}-1})$$ 
%$(X^p_{t-\gamma_{pq}: t-\gamma_{pq}+\lambda_{pq}}, (X^{r_i}_{t-\Gamma_{pq|r_i}})_{1\leq i \leq K})$, $(X^q_{t:t+\lambda_{pq}}, (X^{r_i}_{t-\Gamma_{pq|r_i}})_{1\leq i \leq K})$ and $(X^{r_i}_{t-\Gamma_{pq|r_i}})_{1\leq i \leq K}$ 
respectively, and $n_{\gamma_{pq}, \gamma_{qp}}$ the number of observations.
The estimate of the causal temporal mutual information is then given by: 
\begin{align*}
%\label{eq:TMI-estimated}
\widehat{{CTMI}}(X^p ; X^q) = \psi(k) + \frac{1}{n_{ \gamma_{pq}, \gamma_{qp}}}\sum_{i=1}^{n_{ \gamma_{pq}, \gamma_{qp}}} \psi(n^{3}_i) -\psi(n^{1, 3}_i) - \psi(n^{2, 3}_i)
\end{align*}
where $\psi$ denotes the digamma function.

Similarly, for the estimation of the conditional causal temporal mutual information, we denote by $ \epsilon_{ik}/2$ the distance from 
$$(X^{(p;\bar\lambda_{pq})}_t, X^{(q;\bar\lambda_{pq})}_{t+\bar\gamma_{pq}}, X^{p}_{t-1}, X^{q}_{t+\bar\gamma_{pq}-1}, (X^{(r_k;\bar\lambda_k)}_{t - \bar\Gamma_{k}})_{1\leq k \leq K})$$
to its $k$-th neighbor, 
$n_{i}^{1, 3}$, $n_{i}^{2,3}$, and $n_{i}^3$ being the numbers of points with a distance strictly smaller than $\epsilon_{ik}/2$ in the subspace 
\begin{align*}
(X^{(p;\bar\lambda_{pq})}_t, X^{p}_{t-1}, X^{q}_{t+\bar\gamma_{pq}-1}, (X^{(r_k;\bar\lambda_k)}_{t - \bar\Gamma_{k}})_{1\leq k \leq K})&,
(X^{(q;\bar\lambda_{pq})}_{t+\bar\gamma_{pq}}, X^{p}_{t-1}, X^{q}_{t+\bar\gamma_{pq}-1}, (X^{(r_k;\bar\lambda_k)}_{t - \bar\Gamma_{k}})_{1\leq k \leq K}),\\
&\text{ and }
(X^{p}_{t-1}, X^{q}_{t+\bar\gamma_{pq}-1}, (X^{(r_k;\bar\lambda_k)}_{t - \bar\Gamma_{k}})_{1\leq k \leq K})
\end{align*}
%$(X^p_{t-\gamma_{pq}: t-\gamma_{pq}+\lambda_{pq}}, (X^{r_i}_{t-\Gamma_{pq|r_i}})_{1\leq i \leq K})$, $(X^q_{t:t+\lambda_{pq}}, (X^{r_i}_{t-\Gamma_{pq|r_i}})_{1\leq i \leq K})$ and $(X^{r_i}_{t-\Gamma_{pq|r_i}})_{1\leq i \leq K}$ 
respectively, and $n_{ \gamma_{rp}, \gamma_{rq}}$ the number of observations.
The estimate of the conditional causal temporal mutual information is then given by: 
\begin{align*}
&\widehat{{CTMI}}(X^p ; X^q \mid X^\textbf{R}) = \psi(k) + \frac{1}{n_{ \gamma_{rp}, \gamma_{rq}}}\sum_{i=1}^{n_{ \gamma_{rp}, \gamma_{rq}}} \psi(n^{3}_i) -\psi(n^{1, 3}_i) - \psi(n^{2, 3}_i)
\end{align*}
where $\psi$ denotes the digamma function.

To detect independencies through CTMI, we rely on the following permutation test: 
\begin{Definition}[Permutation test of CTMI]\label{def:ctmi_pvalue}
	Given $X^p$, $X^q$, and $X^{\textbf{R}}$, the p-value associated with the permutation test of CTMI is given by:
	\begin{equation}
	p = \frac{1}{B}\sum_{b=1}^B \mathds{1}_{\widehat{{CTMI}}((X^p)_b ; X^q \mid X^\textbf{R})\ge \widehat{{CTMI}}(X^p ; X^q \mid X^\textbf{R})}, \label{eq:ctmi_pvalue}
	\end{equation}
	where $(X^p)_b$ is a permuted version of $X^p$ and follows the local permutation scheme presented in~\cite{Runge_2018}. 
\end{Definition}
The advantage of the scheme presented in \cite{Runge_2018} is that it preserves marginals by drawing as much as possible without replacement and it performs local permutation, which ensures that by permuting $X^p$, the dependence between $X^p$ and $X^r$ is not destroyed.

Note that Definition~\ref{def:ctmi_pvalue} is applicable to the causal temporal mutual information (when $\textbf{R}$ is empty) and to the conditional causal temporal mutual information.

\subsection{Extension to Time Series with Different Sampling Rates}

The above development readily applies to time series with different sampling rates, as one can define window-based representations of the two time series, as well as a sequence of joint observations. 

Indeed, as one can note, Definition~\ref{def:chapEntropyBased_complag} does not rely on the fact that the time series have the same sampling rates. % and can directly be used on time series with different sampling rates to define compatible window sizes and lags.
Figure~\ref{fig:sampling-rates} displays two time series $X^{p}$ and $X^{q}$ with different sampling rates, where, while $\lambda_{pq}=2$ and $\lambda_{qp}=3$, the time spanned by each window is the same. 
The joint sequence of observations, relating pairs of windows from $X^{p}$ and $X^{q}$ in the form {\small{$S=\{(X^{(p;\lambda_{pq})}_{1_p}, X^{(q;\lambda_{qp})}_{1_q}), \cdots,(X^{(p;\lambda_{pq})}_{n_p}, X^{(q;\lambda_{qp})}_{n_q})\}$}}, should, however, be such that, for all indices $i$ of the sequence, one has: {\small{$s(X^{(q;\lambda_{qp})}_{i_q}) = s(X^{(p;\lambda_{pq})}_{i_p}) + \gamma_{pq}$}}, where $s(X)$ represents the starting time of the window $X$, and $\gamma_{pq}$ is constant over time. This is not the case for the first example, but is true for the second one, which is a relevant sequence of observations.
\begin{figure}[H]
	\centering
	\begin{subfigure}{0.45\textwidth}
		\centering
		\begin{tikzpicture}[font=\footnotesize, yscale=0.85, xscale=0.65]
		\renewcommand{\axisdefaulttryminticks}{4}
		\pgfplotsset{every major grid/.append style={densely dashed}}
		\begin{axis}[
		axis lines=none,
		height = 4.5cm, width =1.8\textwidth,
		ymin=0.0,
		xticklabels=\empty,
		ymax=3,
		ytick={1,2},
		grid=major,
		yticklabels={$\mathbf{X}^{p}$,$\mathbf{X}^{q}$},
		scaled ticks=true,
		]
		\addplot[black,only marks,mark=*] plot coordinates{
			(1, 1) (2, 1) (3, 1) (4, 1) (5, 1)
		};
		\addplot[black,only marks, mark=*] plot coordinates{
			(1, 2) (3, 2) (5, 2)
		};
		\draw [black, thick, decorate,decoration={brace,amplitude=5pt,mirror,raise=4ex}] (-0.2cm,1.53cm) -- (3.75cm,1.53cm) node[midway,yshift=-3.3em, xshift=-1.9em]{$X_{t_p}^{(p;3)}$};
		\draw [black, thick, decorate,decoration={brace,amplitude=5pt}] (1.55cm,1.02cm) -- (5.5cm,1.02cm) node[midway,yshift=1.2em, xshift=1.8em]{$X_{t_p+1}^{(p;3)}$};
		\draw [black, thick, decorate,decoration={brace,amplitude=5pt,mirror,raise=4ex}] (-0.2cm,2.50cm) -- (3.75cm,2.50cm) node[midway,yshift=-3.3em, xshift=-1.9em]{$X_{t_q}^{(q;2)}$};
		\draw [black, thick, decorate,decoration={brace,amplitude=5pt}] (3.35cm,1.99cm) -- (7.3cm,1.99cm) node[midway,yshift=1.2em, xshift=1.8em]{$X_{t_q+1}^{(q;2)}$};
		\node[anchor=west] (source) at (-1.0cm,1.95cm){};
		\node (destination) at (7.8cm,1.9cm){};
		\draw[->](source)--(destination);
		\node[anchor=west] (source) at (-1.0cm,0.95cm){};
		\node (destination) at (7.8cm,0.95cm){};
		\draw[->](source)--(destination);
		\draw[black] (-0.5cm, 2.3cm) node[right]{Incompatible lags};
		\end{axis}
		\end{tikzpicture}
		\caption{}
		\label{fig:sampling-rates-a}
	\end{subfigure}
	%	\vspace{-0.3cm}
	\hfill
	\begin{subfigure}{0.45\textwidth}
		\centering
		\begin{tikzpicture}[font=\footnotesize, yscale=0.85, xscale=0.65]
		\pgfplotsset{every major grid/.append style={densely dashed}}
		%\pgfplotsset{every axis legend/.append style={cells={anchor=west},fill=white, at={(0.02,0.98)}, anchor=north west}}
		\begin{axis}[
		axis lines=none,
		height = 4.5cm, width = 1.8\textwidth,
		ymin=0.0,
		xticklabels=\empty,
		ymax=3,
		ytick={1,2},
		yticklabels={$\mathbf{X}^{p}$,$\mathbf{X}^{q}$},
		scaled ticks=true,
		]
		\addplot[black,only marks,mark=*] plot coordinates{
			(1, 1) (2, 1) (3, 1) (4, 1) (5, 1)
		};
		
		\addplot[black,only marks, mark=*] plot coordinates{
			(1, 2) (3, 2) (5, 2)
		};
		
		\draw [black, thick, decorate,decoration={brace,amplitude=5pt,mirror,raise=4ex}] (-0.2cm,1.53cm) -- (3.75cm,1.53cm) node[midway,yshift=-3.3em, xshift=-1.9em]{$X_{t_p}^{(p;3)}$};
		\draw [black, thick, decorate,decoration={brace,amplitude=5pt}] (3.35cm,1.02cm) -- (7.3cm,1.02cm) node[midway,yshift=1.2em, xshift=1.8em]{$X_{(t+1)_p}^{(p;3)}$};

		\draw [black, thick, decorate,decoration={brace,amplitude=5pt,mirror,raise=4ex}] (-0.2cm,2.50cm) -- (3.75cm,2.50cm) node[midway,yshift=-3.3em, xshift=-1.9em]{$X_{t_q}^{(q;2)}$};
		\draw [black, thick, decorate,decoration={brace,amplitude=5pt}] (3.35cm,1.99cm) -- (7.3cm,1.99cm) node[midway,yshift=1.2em, xshift=1.8em]{$X_{(t+1)_q}^{(q;2)}$};
		\node[anchor=west] (source) at (-1.0cm,1.95cm){};
		\node (destination) at (7.8cm,1.9cm){};
		\draw[->](source)--(destination);
		\node[anchor=west] (source) at (-1.0cm,0.95cm){};
		\node (destination) at (7.8cm,0.95cm){};
		\draw[->](source)--(destination);
		\draw[black] (-0.5cm, 2.3cm) node[right]{Compatible lags};
		\end{axis}
		\end{tikzpicture}
		\caption{}
		\label{fig:sampling-rates-b}
	\end{subfigure}
	\caption{Illustration for constructing sequences of windows for two time series with different sampling rates. In (\textbf{a}) the construction represents incompatible lags and in (\textbf{b}) the construction represents compatible lags.}
	\label{fig:sampling-rates}
\end{figure}

If the two time series are sufficiently long, there always exists a correct sequence of joint observations. Indeed, if the window sizes $\lambda_{pq}$ and $\lambda_{qp}$ are known, let {\small{$\gamma_{pq} = s(X^{(q;\lambda_{qp})}_{1}) - s(X^{(p;\lambda_{pq})}_{1})$}}. Furthermore, let $N_p$ and $N_q$ denote the number of observations per time unit %where the time unit
({the time unit corresponds to the largest (integer) time interval according to the sampling rates of the different time series. For example, if a time series has a sampling rate of}%MDPI: The footnote is not allowed, moved into the text, please confirm. (Confirmed)
~$10$ per second and another a sampling rate of $3$ per $10$ min, then the time unit is equal to $10$ min). Then, $\lambda_{pq}$, $\lambda_{qp}$, and $\gamma_{pq}$ are compatible through the set of joint observations $S$ 
with 
%\begin{align*}
{\small{$s(X^{(p;\lambda_{pq})}_{i_p}) = s(X^{(p;\lambda_{pq})}_{1}) + (i_p-1) LCM(N_p,N_q)$}}
%\end{align*}
and
%similarly for $w^{(q;\lambda_{qp})}_{i_q}$,
%\begin{align*}
{\small{$s(X^{(q;\lambda_{qp})}_{i_q}) = s(X^{(q;\lambda_{qp})}_{1}) + (i_q-1) LCM(N_p,N_q),$}}
%\end{align*}
such that $LCM$ is the lowest common multiple. %$i$ varies from $1$ till the length of $S$. 

Note that this methodology for handling different sampling rates is not unique to our proposal and can be used as soon as lags and windows are defined.

\section{Causal Discovery Based on Causal Temporal Mutual Information}\label{sec:chapEntropyBased_cd}

We present in this section two new methods for causal discovery in time series based on the causal temporal mutual information introduced above to construct the skeleton of the causal graph. The first method assumes causal sufficiency (there exists no hidden common cause of two observable time series), while the second method is an extension of the first for the case where causal sufficiency is not satisfied.
In both methods, the skeleton is oriented on the basis of the entropy reduction principle in addition to classical constraint-based orientation rules (the rules used in the PC algorithm or the rules used in the FCI algorithm). Our methods assume both the causal Markov condition and faithfulness of the data distribution, which are classical assumptions for causal discovery within constraint-based methods in addition to acyclicity in the summary causal graph. 

\subsection{Without Hidden Common Causes}
We present here our main method, which assumes causal sufficiency. We start by describing how we infer the skeleton of the summary graph, then we show how we orient edges using classical PC-rules in addition to rules based on the entropy reduction principle.

\subsubsection{Skeleton Construction}
\label{sec:chapEntropyBased_skel}

We follow the same steps as the ones of the PC algorithm \citep{Spirtes_2000}, which assumes that all variables are observed. It aims at building causal graphs by orienting a skeleton obtained, from a complete graph, by removing edges connecting independent variables. The summary causal graphs considered are directed acyclic graphs (DAGs) in which self-loops are allowed to represent temporal dependencies within a time series.

Starting with a complete graph that relates all time series, the first step consists of computing CTMI for all pairs of time series and removing edges if the two time series are considered independent. Once this is done, one checks, for the remaining edges, whether the two time series are conditionally independent (the edge is removed) or not (the edge is kept). Starting from a single time series connected to $X^{p}$ or $X^{q}$, the set of conditioning time series is gradually increased until either the edge between $X^{p}$ and $X^{q}$ is removed or all time series connected to $X^{p}$ and $X^{q}$ have been considered. We will denote by $\text{Sepset}(p,q)$ the separation set of $X^{p}$ and $X^{q}$, which corresponds to the smallest set of time series connected to $X^{p}$ and $X^{q}$ such that $X^{p}$ and $X^{q}$ are conditionally independent given this set. Note that we make use of the same strategy as the one used in PC-stable \citep{colombo}, which consists of sorting time series according to their $\text{CTMI}$ scores and, when an independence is detected, removing all other occurrences of the time series. This leads to an order-independent procedure.
The following theorem states that the skeleton obtained by the above procedure is the true one.
\begin{Theorem}
	\label{them:s}
	Let $\mathcal{G} = (V, E)$ be a summary causal graph, and assume that we are given perfect conditional independence information about all pairs of variables $(X^{p},X^{q})$ in $V$ given subsets $S \subseteq V\backslash\{X^{p},X^{q}\}$. Then, the skeleton previously constructed is the skeleton of $\mathcal{G}$.
\end{Theorem}

\begin{proof}
	Let us consider two time series $X^{p}$ and $X^{q}$. If they are independent given $X^{\textbf{R}}$, then $\text{CTMI}(X^{p};X^{q}|X^{\textbf{R}})=0$, as otherwise, the conditional mutual information between $X^{p}$ and $X^{q}$ would be non-null and the two time series would not be conditionally independent as we are given perfect information. By the causal Markov and faithfulness conditions, there is no edge in this case between $X^{p}$ and $X^{q}$ in the corresponding skeleton, as in the true one. Conversely, if $\text{CTMI}(X^{p};X^{q}|X^{\textbf{R}})=0$ for any $X^{\textbf{R}}$, then the two time series cannot be dependent conditioned on $X^{\textbf{R}}$. Indeed, if this were the case, as we are given perfect conditional information, there would exist a lag $\gamma$ and two window sizes $\lambda_{pq}$ and $\lambda_{pq}$ such that $I(X^{(p;\lambda_{pq})}_t;X^{(q;\lambda_{qp})}_{t+\gamma}|X^{\textbf{R}}) > 0$ with $0 < \lambda_{pq}, \lambda_{qp} \le \gamma$. In this case, the two windows of size $\lambda_{max}$ centered on time point $t$ in both $X^p$ and $X^q$ contain the windows of sizes $\lambda_{pq}$ and $\lambda_{pq}$ separated by a lag $\gamma$ in $X^{p}$ and $X^{q}$ as $\lambda_{max}=2\gamma_{max}+1$. Thus, $\text{CTMI}(X^{p};X^{q}|X^{\textbf{R}})$ would be positive as this quantity cannot be less than $I(X^{(p;\lambda_{pq})}_t;X^{(q;\lambda_{qp})}_{t+\gamma}|X^{\textbf{R}})$, which leads to a contradiction. Finally, as we tested all necessary conditioning sets in the construction of the skeleton, we have the guarantee of removing all unnecessary edges.
\end{proof}

\subsubsection{Orientation}
\label{sec:PC}

Once the skeleton has been constructed, one tries to orient as many edges as possible using the standard PC-rules \citep{Spirtes_2000}, which yields a completed partially directed acyclic graph (CPDAG).

\begin{PC-rule}[Origin of causality]
	In an unshielded triple $X^{p} - X^{r} - X^{q}$, if $X^{r} \notin \texttt{sepset}(p,q)$, then $X^{r}$ is an unshielded collider: $X^{p} \rightarrow X^{r} \leftarrow X^{q}$.
	\label{prop:oc}
\end{PC-rule}
\begin{PC-rule}[Propagation of causality]
	In an unshielded triple $X^{p} \rightarrow X^{r} - X^{q}$, if $X^{r} \in \texttt{sepset}(p,q)$,
	then orient the unshielded triple as $X^{p} \rightarrow X^{r} \rightarrow X^{q}$.
	\label{prop:pc}
\end{PC-rule}
\begin{PC-rule}
	If there exist a direct path from $X^{p}$ to $X^{q}$ and an edge between $X^{p}$ and $X^{q}$, then orient $X^{p} \rightarrow X^{q}$.
	\label{prop:r2}
\end{PC-rule}
\begin{PC-rule}
	If there exists an unshielded triple $X^{p} \rightarrow X^{r} \leftarrow X^{q}$ and an unshielded triple $X^{p} - X^{s} - X^{q}$, then orient $X^{s} \rightarrow X^{r}$.
	\label{prop:r3}
\end{PC-rule}

As we are using here the standard PC-rules, we have the following theorem, the proof of which directly derives from the results on PC \citep{Spirtes_2000}.
\begin{Theorem}[Theorem 5.1 of \cite{Spirtes_2000}]
	\label{them:cpdag}
	Let the distribution of $V$ be faithful to a DAG $\mathcal{G}=(V,E)$, and assume that we are given perfect conditional independence information about all pairs of variables $(X^p, X^q)$ in $V$ given subsets $X^{\textbf{R}} \subseteq V\backslash\{X^p, X^q\}$. Then, the skeleton constructed previously followed by PC-rules~\ref{prop:oc}, \ref{prop:pc}, \ref{prop:r2}, and \ref{prop:r3} represents the CPDAG of $\mathcal{G}$. 
\end{Theorem}

In addition to the PC orientation rules, we introduce two new rules, which are based on the notion of \textit{possible spurious correlations} and the mutual information we have introduced. The notion of possible spurious correlations captures the fact that two variables may be correlated through relations that do not only correspond to direct causal relations between them. It is formalized as follows:
\begin{Definition}[Possible spurious correlations]
	\label{def:possible_common_ancestors}
	We say that two nodes $X^p-X^q$ have possible spurious correlations if there exists a path between them that neither contains the edge $X^p-X^q$ nor any collider.
\end{Definition}
Interestingly, when two connected variables do not have possible spurious correlations, one can conclude about their orientation using CTMI.
\begin{Property}
	\label{coro}
	Let us assume that we are given perfect conditional independence information about all pairs of variables $(X^{p},X^{q})$ in $V$ given subsets $S \subseteq V\backslash\{X^{p},X^{q}\}$. Then, every non-oriented edge in the CPDAG obtained by the above procedure corresponds to a {prima facie} cause and by, causal sufficiency, to a true causal relation between the related time series. Furthermore, the orientation of an unoriented edge between two nodes $X^p$ and $X^q$ that do not have possible spurious correlations is given by the ``direction'' of the optimal lag in $CTMI(X^p,X^q)$, assuming that the maximal window size is larger than the longest lag $\gamma_{max}$ between causes and effects.
\end{Property}

%Proof in Appendix~\ref{appendix:proof}.
%
\begin{proof}
	The first part of {Property}~\ref{coro} directly derives from {Property}~\ref{prop:entropic_prima_facie_cause}. As we assume that we are given perfect conditional information, the skeleton is the true one from Theorem~\ref{them:s}. Thus, if two variables do not have possible spurious correlations, the only correlations observed between them correspond to a causal relation. We now need to prove that the optimal lag can be used to orient edges between any pair of variables $X^{p}$ and $X^{q}$.%MDPI: please confirm if need change to Property. (Changed to Property)
	
	Without loss of generality, let us assume that $X^{p}$ causes $X^{q}_t$, for any time $t$, via the $K$ time instants $\{t-\gamma, t-\gamma_1, \cdots, t-\gamma_{K-1}\}$ with $0 < \gamma_{K-1} < \cdots < \gamma_{1} < \gamma$. First, let us consider a window of size 1 in $X^q$ and a window of arbitrary size $\lambda$ in $X^{p}$ with a lag $\gamma_{pq}$ set to $\gamma' \ge 0$. As $\gamma' \ge 0$, there is no cause of $X^q_t$ in the window $X^{(p;\lambda)}_{t+\gamma'}$. Furthermore, the only observed correlations between $X^{p}$ and $X^{q}$ correspond to causal relations. We thus have:
	\[
	I(X^{(q;1)}_t;X^{(p;\lambda)}_{t+\gamma'}|X^{(q;1)}_{t-1},X^{(p;1)}_{t+\gamma'-1}) = 0,
	\]
	as $X^{q}_t$ and all variables in $X^{(p;\lambda)}_{t+\gamma'}$ are independent of each other. One the contrary, for the same window size in $X^p$ and a lag $\gamma_{pq}$ set to $-\gamma$ with $\gamma > 0$, one has:
	\[
	I(X^{(q;1)}_t;X^{(p;\gamma)}_{t-\gamma}|X^{(q;1)}_{t-1},X^{(p;1)}_{t-\gamma-1}) \ge I(X^{(q;1)}_t;X^{(p;1)}_{t-\gamma}|X^{(q;1)}_{t-1},X^{(p;1)}_{t-\gamma-1}) > 0.
	\]
	The first inequality derives from Inequality~\ref{ineq:ctmi}. The second inequality is due to the fact that $X^{p}_{t-\gamma}$ is a true cause of $X^{q}_t$ and the fact that we are given perfect information. Thus, when considering a window of size $1$ for $X^q$, the optimal lag given by CTMI will necessarily go from $X^p$ to $X^q$, which corresponds to the correct orientation.
	
	We now consider the case where we have a window of arbitrary size $\lambda'$ in $X^q$. Let us further consider a window of arbitrary size $\lambda$ in $X^{p}$ with a lag $\gamma_{pq}$ set to $\gamma' \ge 0$. If $\lambda' < \gamma'+\gamma_{K-1}$, there is no causal relations between the elements in $X^{(q;\lambda')}_t$ and the elements in $X^{(p;\lambda)}_{t+\gamma'}$ and the mutual information between these two windows is $0$. Otherwise, one can decompose this mutual information as:
	%
%	\begin{adjustwidth}{-\linewidth}{0cm}\vspace{-12pt}
		%\centering %% If there is a figure in wide page, please release command \centering
		\begin{align*}
		&I(X^{(q;\lambda')}_t;X^{(p;\lambda)}_{t+\gamma'}|X^{(q;1)}_{t-1},X^{(p;1)}_{t+\gamma'-1}) \\
		& = I(X^{(q;\gamma'+\gamma_{K-1})}_t;X^{(p;\lambda)}_{t+\gamma'}|X^{(q;1)}_{t-1},X^{(p;1)}_{t+\gamma'-1}) 
		+ I(X^{(q;\lambda'-\gamma'-\gamma_{K-1})}_{t+\gamma'+\gamma_{K-1}};X^{(p;\lambda)}_{t+\gamma'}|X^{(q;1)}_{t+\gamma'+\gamma_{K-1}-1},X^{(p;1)}_{t+\gamma'-1}),
		\end{align*}
%	\end{adjustwidth}
	%
	as the conditioning on $X^{(q;\gamma'+\gamma_{K-1})}_t$ and $X^{(q;1)}_{t-1}$ amounts to conditioning on the instant $X^{(q;1)}_{t+\gamma'+\gamma_{K-1}-1}$ due to the first-order Markov self-causal assumption.
	
	As there are no causal relations between the elements in $X^{(q;\gamma'+\gamma_{K-1})}_t$ and the elements in $X^{(p;\lambda)}_{t+\gamma'}$, the first term on the right-hand side is $0$. Using a similar decomposition in order to exclude elements at the end of $X^{(p;\lambda)}_{t+\gamma'}$, which do not cause any element in $X^{(q;\lambda')}_t$, one obtains:
	\begin{align*}
	&I(X^{(q;\lambda')}_t;X^{(p;\lambda)}_{t+\gamma'}|X^{(q;1)}_{t-1},X^{(p;1)}_{t+\gamma'-1}) \\
	&= I(X^{(q;\lambda'-\gamma'-\gamma_{K-1})}_{t+\gamma'+\gamma_{K-1}};X^{(p;\min(\lambda,\lambda'-\gamma_{K-1}-\gamma'))}_{t+\gamma'}|X^{(q;1)}_{t+\gamma'+\gamma_{K-1}-1},X^{(p;1)}_{t+\gamma'-1}).
	\end{align*}
	Let us now consider the window in $X^p$ of size $\lambda'$ with a lag $\gamma_{pq}$ set to $-\gamma_{K-1}$. Using the same reasoning as before, one obtains:
	\begin{align}\label{eq:decomp}
	I(X^{(q;\lambda')}_t;X^{(p;\lambda')}_{t-\gamma_{K-1}}|X^{(q;1)}_{t-1},X^{(p;1)}_{t-\gamma_{K-1}-1}) &= I(X^{(q;\gamma'+\gamma_{K-1})}_t;X^{(p;\lambda')}_{t-\gamma_{K-1}}|X^{(q;1)}_{t-1},X^{(p;1)}_{t-\gamma_{K-1}-1}) \nonumber \\
	& \hspace{-1cm} + I(X^{(q;\lambda'-\gamma'-\gamma_{K-1})}_{t+\gamma'+\gamma_{K-1}};X^{(p;\lambda')}_{t-\gamma_{K-1}}|X^{(q;1)}_{t+\gamma'+\gamma_{K-1}-1},X^{(p;1)}_{t-\gamma_{K-1}-1}),
	\end{align}
	with:
	\begin{align*}
	&I(X^{(q;\lambda'-\gamma'-\gamma_{K-1})}_{t+\gamma'+\gamma_{K-1}};X^{(p;\lambda')}_{t-\gamma_{K-1}}|X^{(q;1)}_{t+\gamma'+\gamma_{K-1}-1},X^{(p;1)}_{t-\gamma_{K-1}-1}) \\
	&\ge I(X^{(q;\lambda'-\gamma'-\gamma_{K-1})}_{t+\gamma'+\gamma_{K-1}};X^{(p;\min(\lambda,\lambda'-\gamma_{K-1}-\gamma'))}_{t+\gamma'}|X^{(q;1)}_{t+\gamma'+\gamma_{K-1}-1},X^{(p;1)}_{t+\gamma'-1}),
	\end{align*}
	as the window $X^{(p;\lambda')}_{t-\gamma_{K-1}}$ contains the window $X^{(p;\min(\lambda,\lambda'-\gamma_{K-1}-\gamma'))}_{t+\gamma'}$. %and the conditioning on the past is the same for each time instant. 
	In addition, the first term on the right-hand side of Equation~(\ref{eq:decomp}) is strictly positive as all the elements in $X^{(q;\gamma'+\gamma_{K-1})}_t$ have causal relations in $X^{(p;\lambda')}_{t-\gamma_{K-1}}$. Thus, the mutual information obtained with the negative lag $-\gamma_{K-1}$ is better than the one obtained with any positive lag: 
	
	\[
	I(X^{(q;\lambda')}_t;X^{(p;\lambda')}_{t-\gamma_{K-1}}|X^{(q;1)}_{t-1},X^{(p;1)}_{t-\gamma_{K-1}-1})
	> I(X^{(q;\lambda')}_t;X^{(p;\lambda)}_{t+\gamma'}|X^{(q;1)}_{t-1},X^{(p;1)}_{t+\gamma'-1}); \]
	meaning that the optimal lag given by CTMI will necessarily go from $X^p$ to $X^q$, which corresponds to the correct orientation.
\end{proof}

The following orientation rule is a direct application of the above property.
\begin{ER-rule}[Entropy reduction---$\gamma$ ]
	In a pair $X^{p} - X^{q}$, such $X^{p}$ and $X^{q}$ do not have a possible spurious correlation,
	if $\bar\gamma_{pq} > 0$, then orient the edge as: $X^{p} \rightarrow X^{q}$.
	\label{prop:ER0}
\end{ER-rule}
%
%Furthermore, we make use of the following heuristic rule to orient additional edges when the optimal lag $\bar\gamma_{pq}$ is null. This heuristic is based on the intuition that the difference in the window sizes is, when the optimal lag is null, an indication of the direction of the causal relation when there are common causes with different lags for the two time series under consideration.
Furthermore, we make use of the following rule to orient additional edges when the optimal lag $\bar\gamma_{pq}$ is null based on the fact that CTMI increases asymmetrically with respect to the increase of $\lambda_{pq}$ and $\lambda_{qp}$ (Figure~\ref{fig:ctmi_asymetric}). This rule infers the direction of the cause by checking the difference in the window sizes as the window size of the cause cannot be greater than the window size of the effect. 

\begin{ER-rule}[Entropy reduction---$\lambda$]
	In a pair $X^{p} - X^{q}$, such $X^{p}$ and $X^{q}$ do not have a possible spurious correlation,
	if $\bar\gamma_{pq} = 0$ and $\bar\lambda_{pq}<\bar\lambda_{qp}$, 
	%and for all $X^{r}$ s.t. $X^{q} \leftarrow X^{r} \rightarrow X^{p}$, $\bar\gamma_{rp}, \bar\gamma_{rq}<\bar\lambda_{pq}$,
	then orient the edge as: $X^{p} \rightarrow X^{q}$. 
	\label{prop:ER1}
\end{ER-rule}

%\begin{property}
%	The above rules permits us to orient all lagged causal relations except sheilded triples and all instantaneous relations 
%\end{property}
%
%\begin{proof}
%	For a pair $X^{p} - X^{q}$ in a CPDAG, such $X^{p}$ and $X^{q}$ do not have a possible common ancestor, if $\bar\gamma_{pq} = 0$, then, by there exist at least two entropic prima facie which can be genuine or spurious from a causal chain. If $\bar\gamma_{pq} = 0$ reflect the genuine causal relation and $\bar\lambda_{pq}$ ...
%	If $\bar\gamma_{pq} = 0$ reflect the spurious causal relation from a causal chain and $\bar\lambda_{pq}$ ...
%\end{proof}

%For partially oriented shielded triples we can do better: if $X^{r} \leftarrow X^{p} - X^{q} \rightarrow X^{r} -$, apply ER-rules on $X^{p} - X^{q}$; if $X^{r} \rightarrow X^{p} - X^{q} \leftarrow X^{r} -$, apply compute $\bar\gamma_{pq}, \bar\lambda_{pq}, \bar\lambda_{qp} = arg\max_{(\lambda_{pq}, \lambda_{qp}, \gamma_{pq}) \in \mathcal{C}^{(p,q)}} I(X_t^{(p;\lambda_{pq})};X_{t +\gamma_{pq}}^{(q;\lambda_{qp})} |X^{(p;1)}_{t-1},X_{t +\gamma_{pq}-1}^{(q;1)} )$ then apply ER-rules on $\bar\gamma_{pq}, \bar\lambda_{pq}, \bar\lambda_{qp}$.

In practice, we also apply the ER-rule~\ref{prop:ER0} before PC-rules, because, experimentally, we found that the ER-rule~\ref{prop:ER0} is more reliable than the PC-rule~\ref{prop:oc} in detecting lagged unshielded colliders, especially in the case of a low sample size.

%Finally, given the graph $\mathcal{G}$ inferred with the above procedure, one can verify for each node $X^q$ in $\mathcal{G}$ if it is self causal by checking if for all $t$, $CTMI(X^q_t ; X^q_{t-1} \mid \text{Par}(X^q_{t}))$ in $\mathcal{G}$.

%We call our method \texttt{PCTMI}, a pseudo-code in available in Appendix~\ref{appendix:algo_complex}.

We call our method \texttt{PCTMI}; the pseudo-code is available in Algorithm~\ref{algo:PCTMI}. $\text{Adj}(X^{q}, \mathcal{G})$ represents all adjacent nodes to $X^{q}$ in the graph $\mathcal{G}$, and $\texttt{sepset}(p,q)$ is the separation set of $X^{p}$ and $X^{q}$. The output of the algorithm is a CPDAG version of the summary graph such that all lagged relations are oriented, but instantaneous relations are partially oriented.

\begin{algorithm}[H]
	\caption{\texttt{PCTMI}.}
	\label{algo:PCTMI}
	\begin{algorithmic}
		\STATE $X$ a $d$-dimensional time series of length $T$, $\gamma_{\max}\in \mathbb{N}$ the maximum number of lags, $\alpha$ a significance threshold\\	
		\STATE Form a complete undirected graph $\mathcal{G}=(V,E)$ with $d$ nodes
		\STATE n = 0
		\WHILE{there exists $X^{q} \in V$ such that $\text{card}(\text{Adj}(X^{q}, \mathcal{G})) \ge n+1$}
		\STATE $\textbf{D} = list()$
		\FOR{$X^{q} \in V$ s.t. $\text{card}(\text{Adj}(X^{q}, \mathcal{G})) \ge n+1$}
		\FOR{$X^{p} \in \text{Adj}(X^{q}, \mathcal{G})$} 
		\FOR{all subsets $X^{\mathbf{R}} \subset \text{Adj}(X^{q}, \mathcal{G})\setminus \{X^{p}\}$ such that $\text{card}(X^{\mathbf{R}} )=n$ and ($\Gamma_{rp}\ge 0$ or $\Gamma_{rq}\ge 0 $) for all $r \in \mathbf{R}$}
		%		all pairs of adjacent vertices $(X^{p},X^{q})$
		\STATE $y_{q,p,\textbf{R}} = \text{CTMI}(X^{p};X^{q} \mid X^{\mathbf{R}} )$
		\STATE append $(\textbf{D}, \{X^{q},X^{p}, X^{\mathbf{R}}, y_{q,p,\textbf{R}} \}))$
		\ENDFOR 
		\ENDFOR
		\ENDFOR 
		\STATE Sort $\textbf{D}$ by increasing order of $y$
		\WHILE {$\textbf{D}$ is not empty}
		\STATE $\{X^{q},X^{p},X^{\mathbf{R}},y\} = \text{pop}(\textbf{D})$
		\IF{$X^{p} \in \text{Adj}(X^{q}, \mathcal{G})$ and $X^{\mathbf{R}} \subset \text{Adj}(X^{q}, \mathcal{G})$}
		\STATE Compute $z$ the p-value of $\text{CTMI}(X^{p};X^{q} \mid X^{\mathbf{R}} )$ given by Equation \eqref{eq:ctmi_pvalue} 
		\IF{test $z>\alpha$}
		\STATE Remove edge $X^{p} - X^{q}$ from $\mathcal{G}$
		\STATE $\text{Sepset}(p,q) = \text{Sepset}(q,p) = X^{\mathbf{R}}$
		\ENDIF
		\ENDIF
		\ENDWHILE
		\STATE n=n+1
		\ENDWHILE		
		
		%		\STATE \textbf{for} each connected pair in $\mathcal{G}$ \textbf{do} apply ER-rules \ref{prop:ER0}
		\STATE \textbf{for} each triple in $\mathcal{G}$, \textbf{do} apply PC-rule \ref{prop:oc}		
		\WHILE{no more edges can be oriented}
		\STATE \textbf{for} each triple in $\mathcal{G}$, \textbf{do} apply PC-rules \ref{prop:pc}, \ref{prop:r2}, and \ref{prop:r3}
		\ENDWHILE
		% 		\FOR{$X^{q} \in V$}
		%   \STATE $z_{q,q} =$ test $\text{CTMI}(X^{q};X^{q} \mid \text{Par}(X^{q} )$
		%   \IF{$z_{q,q}>\alpha$}
		%   \STATE Remove self-loop on $X^{q}$ from $\mathcal{G}$
		%   \ENDIF
		%  \ENDFOR 
		\STATE \textbf{for} each connected pair in $\mathcal{G}$ \textbf{do} apply ER-rules \ref{prop:ER0} and \ref{prop:ER1}
		\STATE \textbf{Return} $\mathcal{G}$ 
	\end{algorithmic}
\end{algorithm}

%, we also test a variant of \texttt{PCTMI} where we drop the possible spurious correlation condition in the ER-rules. We refer to this \texttt{oPCTMI}.

%\begin{theorem}\label{them:pctmi}
%	The orientation rules for \texttt{PCTMI} are sound, meaning that, given a perfect oracle on conditional independence, the CPDAG obtained by these rules is in the Markov equivalence class of the true causal graph.
%\end{theorem}
%Proof in Appendix~\ref{appendix:proof}.

%\begin{algorithm}
%	\caption{PCTMI}
%	\label{algo}
%	\begin{algorithmic}
%		\REQUIRE $\mathcal{X}$ a $d$-dimensional time series
%		\STATE 1. Construction of the skeleton using CTMI %(Section~\ref{sec:skel})
%		\STATE 2. Orient as many edges as possible using \textbf{ER-rules}% 0} and \textbf{ER-rule 1}
%		\STATE 3. Apply orientation rules of PC: (a) orient edges with the origin of causality, then (b) recursively orient new edges with the other rules 
%		\STATE \textbf{Return} the summary graph $\mathcal{G}$ 
%	\end{algorithmic}
%\end{algorithm}

\subsection{Extension to Hidden Common Causes}
When unobserved variables are causing a variable of interest, the PC algorithm is no longer appropriate and one needs to resort to the FCI algorithm introduced in \cite{Spirtes_2000}, which infers a partial ancestral graph (PAG). We extend here the version of this algorithm presented in \cite{Zhang_2008} without the selection bias.

We recall the notations needed for FCI: a double arrow $\rightlefta$ indicates the presence of hidden common causes, while the symbol $\circ$ represents an undetermined mark, and the meta-symbol asterisk * is used as a wildcard, which can denote a tail, an arrow, or a circle. Furthermore, we make use of the notion \textit{Possible-Dsep}, which is defined as follows:
\begin{Definition}[Possible-Dsep \cite{Spirtes_2000}]
	$X^r$ is in the Possible-Dsep set of $X^p$ and $X^q$ if and only if $X^r$ is different from $X^p$ and $X^q$ and there is an undirected path $U$ between $X^p$ and $X^r$ such that, for every subpath $<X^w, X^s, X^v>$ of $U$, either $X^s$ is a collider on the subpath or $X^w$ and $X^v$ are adjacent in the PAG.
\end{Definition}
Lastly, we make use of the following types of paths. A \emph{discriminating path} between $X^{p}$ and $X^{q}$ is a path that includes at least three edges such that each non-endpoint vertex $X^{r}$ on the path is adjacent to $X^{q}$, and $X^{p}$ is not adjacent to $X^{q}$, and every vertex between $X^{p}$ and $X^{r}$ is a collider on the path and is a parent of $X^{q}$. An \emph{uncovered path} is a path where every consecutive triple on the path is unshielded. A \emph{potentially directed path} is a path where the edge between two consecutive vertices is not directed toward the first or against the~second.

From the skeleton obtained in Section~\ref{sec:chapEntropyBased_skel}, 
%as in \texttt{PCTMI}, we start orienting using the ER-rules.
unshielded colliders are detected using the FCI-rule~\ref{fci_prop:oc}.
\begin{FCI-rule}[Origin of causality]
	For each unshielded triple $X^{p} \rightclefts X^{r} \leftcrights X^{q}$, if $X^{r} \notin Sepset(p,q)$, 
	then orient the unshielded triple as a collider: $X^{p} \rightalefts X^{r} \leftarights X^{q}$.
	\label{fci_prop:oc}
\end{FCI-rule}

From this, Possible-Dsep sets can be constructed. As elements of Possible-Dsep sets in a PAG play a role similar to the ones of parents in a DAG, additional edges are removed by conditioning on the elements of the Possible-Dsep sets, using the same strategy as the one given in Section~\ref{sec:chapEntropyBased_skel}. 
All edges are then unoriented, and 
the FCI-rule~\ref{fci_prop:oc} is again applied as some of the edges of the unshielded colliders originally detected may have been removed by the previous step. Then, as in FCI, FCI-rules \ref{fci_prop:r1}, \ref{fci_prop:r2}, \ref{fci_prop:r3}, \ref{fci_prop:r4}, \ref{fci_prop:r8}, \ref{fci_prop:r9}, and \ref{fci_prop:r10} are applied. 

\begin{FCI-rule}
	In an unshielded triple $X^{p} \rightalefts X^{r} \leftcrights X^{q}$, if $X^{r} \in Sepset(p,q)$,
	then orient the unshielded triple as $X^{p} \rightalefts X^{r} \rightalefts X^{q}$.
	\label{fci_prop:r1}
\end{FCI-rule}
\begin{FCI-rule}
	If there exists a triple $X^p \rightarrow X^r \rightalefts X^q$ or a triple $X^p \rightalefts X^r \rightarrow X^q$ with $X^p \rightclefts X^q$, then orient the pair as $X^p \rightalefts X^q$.
	\label{fci_prop:r2}
\end{FCI-rule}
\begin{FCI-rule}
	If there exists an unshielded triple $X^{p} \rightalefts X^{r} \leftarights X^{q}$ and an unshielded triple $X^{p} \rightclefts X^{s} \leftcrights X^{q}$ and $X^{s} \rightclefts X^{r}$, then orient the pair as $X^{s} \rightarrow X^{r}$.
	\label{fci_prop:r3}
\end{FCI-rule}
\begin{FCI-rule}
	If there exists a discriminating path %$u=\langle p, \cdots s,r,q\rangle$ 
	between $X^{p}$ and $X^{q}$ for $X^{r}$ and $X^{r} \leftcrights X^{q}$, then orient $X^{r} \leftcrights X^{q}$ as $X^{r} \rightarrow X^{q}$; otherwise, orient the triple as $X^{s} \rightlefta X^{r} \rightlefta X^{q}$.
	\label{fci_prop:r4}
\end{FCI-rule}

%	\begin{FCI-rule}
%		For every remaining $X^{p} \rightleftc X^{q}$, if there is an uncovered circle path $U = \langle X^p, X^r, \cdots, X^s, X^q\rangle$ between $X^p$ and $X^q$ such that $X^p$ and $X^s$ are not adjacent and $X^q$ and $X^r$ are not adjacent, then orient $X^p \rightleftc X^q$ and every edge on $U$ as undirected edges (-).
%		\label{fci_prop:r5} 
%	\end{FCI-rule}
%	
%	\begin{FCI-rule}
%		If $X^p - X^r \rightclefts X^q$ ($X^p$ and $X^q$ are not necessarily adjacent), then orient the triple as $X^p - X^r \rights X^q$ .
%		\label{fci_prop:r6}
%	\end{FCI-rule}
%	
%	\begin{FCI-rule}
%		If $X^p \rightc X^r \leftcrights X^q$, and $X^p$ and $X^q$ are not adjacent, then orient the triple $X^p \rightc X^r \rights X^q$.
%		\label{fci_prop:r7}
%	\end{FCI-rule}

\begin{FCI-rule8}
	If $X^p \rightarrow X^r \rightarrow X^q$ or $X^p \rightc X^r \rightarrow X^q$ and $X^p \leftcrighta X^q$, then orient $X^p \rightarrow X^q$.
	\label{fci_prop:r8}
\end{FCI-rule8}

\begin{FCI-rule8}
	If $X^p \leftcrighta X^q$ and $U$ is an uncovered potentially directed path from $X^p$ to $X^q$ such that $X^q$ and $X^r$ are not adjacent, then
	orient the pair as $X^p \rightarrow X^q$.
	\label{fci_prop:r9}
\end{FCI-rule8}

\begin{FCI-rule8}
	Suppose $X^p \leftcrighta X^q$, $X^r \rightarrow X^q \leftarrow X^s$, $U_1$ is an uncovered potentially directed path from $X^p$ to $X^r$, and $U_2$ is an uncovered potentially directed path from $X^p$ to
	$X^s$. Let $\mu$ be the vertex adjacent to $X^p$ on $U_1$ ($\mu$ could be $X^r$) and $\omega$ be the vertex adjacent to $X^p$ on $U_2$ ($\omega$ could be $X^s$).
	If $\mu$ and $\omega$ are distinct and are not adjacent, then orient $X^p \leftcrighta X^q$ as $X^p \rightarrow X^q$.
	\label{fci_prop:r10}
\end{FCI-rule8}

Note that we have not included FCI-rules 5, 6, and 7 from \cite{Zhang_2008}, as these rules deal with selection bias, a phenomenon that is not present in the datasets we consider. Including these rules in our framework is nevertheless straightforward.

Finally, as in \texttt{PCTMI}, we orient additional edges using an adapted version of the ER-rules.

\begin{LER-rule}[Latent entropy reduction---$\gamma$ ]
	In a pair $X^{p} \rightclefts X^{q}$, such $X^{p}$ and $X^{q}$ do not have a possible spurious correlation,
	if $\bar\gamma_{pq} > 0$, then orient the edge as: $X^{p} \rightalefts X^{q}$.
	\label{prop:LER0}
\end{LER-rule}

\begin{LER-rule}[Latent entropy reduction---$\lambda$]
	In a pair $X^{p} \rightclefts X^{q}$, such $X^{p}$ and $X^{q}$ do not have a possible spurious correlation,
	if $\bar\gamma_{pq} = 0$ and $\bar\lambda_{pq}<\bar\lambda_{qp}$, 
	%and for all $X^{r}$ s.t. $X^{q} \leftarrow X^{r} \rightarrow X^{p}$, $\bar\gamma_{rp}, \bar\gamma_{rq}<\bar\lambda_{pq}$,
	then orient the edge as: $X^{p} \rightalefts X^{q}$. 
	\label{prop:LER1}
\end{LER-rule}

The overall process, referred to as \texttt{FCITMI}, is described in Algorithm~\ref{algo:FCITMI}. 

\begin{algorithm}[H]
	\caption{\texttt{FCITMI.} }
	\label{algo:FCITMI}
	\begin{algorithmic}
		\REQUIRE $X$ a $d$-dimensional time series of length $T$, $\gamma_{\max}\in \mathbb{N}$ the maximum number of lags, $\alpha$ a significance threshold	
		\STATE Form a complete undirected graph $\mathcal{G}=(V,E)$ with $d$ nodes
		\STATE n = 0
		\WHILE{there exists $X^{q} \in V$ such that $\text{card}(\text{Adj}(X^{q}, \mathcal{G})) \ge n+1$}
		\STATE $\textbf{D} = list()$
		\FOR{$X^{q} \in V$ s.t. $\text{card}(\text{Adj}(X^{q}, \mathcal{G})) \ge n+1$}
		\FOR{$X^{p} \in \text{Adj}(X^{q}, \mathcal{G})$} 
		\FOR{all subsets $X^{\mathbf{R}} \subset \text{Adj}(X^{q}, \mathcal{G})\setminus \{X^{p}\}$ such that $\text{card}(X^{\mathbf{R}} )=n$ and ($\gamma_{rp}\ge 0$ or $\gamma_{rq}\ge 0 $) for all $r \in \mathbf{R}$}
		%		all pairs of adjacent vertices $(X^{p},X^{q})$
		\STATE $y_{q,p,\textbf{R}} = \text{CTMI}(X^{p};X^{q} \mid X^{\mathbf{R}} )$
		\STATE append $(\textbf{D}, \{X^{q},X^{p}, X^{\mathbf{R}}, y_{q,p,\textbf{R}} \}))$
		\ENDFOR 
		\ENDFOR
		\ENDFOR 
		\STATE Sort $\textbf{D}$ by increasing order of $y$
		\WHILE {$\textbf{D}$ is not empty}
		\STATE $\{X^{q},X^{p},X^{\mathbf{R}},y\} = \text{pop}(\textbf{D})$
		\IF{$X^{p} \in \text{Adj}(X^{q}, \mathcal{G})$ and $X^{\mathbf{R}} \subset \text{Adj}(X^{q}, \mathcal{G})$}
		\STATE Compute $z$ the p-value of $\text{CTMI}(X^{p};X^{q} \mid X^{\mathbf{R}} )$ given by Equation \eqref{eq:ctmi_pvalue} 
		\IF{test $z>\alpha$}
		\STATE Remove edge $X^{p} - X^{q}$ from $\mathcal{G}$
		\STATE $\text{Sepset}(p,q) = \text{Sepset}(q,p) = X^{\mathbf{R}}$
		\ENDIF
		\ENDIF
		\ENDWHILE
		\STATE n=n+1
		\ENDWHILE		
		
		%\STATE \textbf{for} each connected pair in $\mathcal{G}$ \textbf{do} apply ER-rules \ref{prop:ER0}, \ref{prop:ER1}
		\STATE \textbf{for} each triple in $\mathcal{G}$, \textbf{do} apply FCI-rule \ref{fci_prop:oc}		
		\STATE using Possible-Dsep sets, remove edges using CTMI
		\STATE Reorient all edges as $\rightleftc$ in $\mathcal{G}$
		%\STATE \textbf{for} each connected pair in $\mathcal{G}$ \textbf{do} apply ER-rules.
		\STATE \textbf{for} each triple in $\mathcal{G}$, \textbf{do} apply FCI-rule \ref{fci_prop:oc}		
		\WHILE{edges can be oriented}
		\STATE \textbf{for} each triple in $\mathcal{G}$, apply FCI-rules \ref{fci_prop:r1}, \ref{fci_prop:r2}, \ref{fci_prop:r3}, \ref{fci_prop:r4}, \ref{fci_prop:r8}, \ref{fci_prop:r9}, and \ref{fci_prop:r10} 
		\STATE \textbf{for} each connected pair in $\mathcal{G}$, \textbf{do} apply ER-rules \ref{prop:LER0} and \ref{prop:LER1}.
		\ENDWHILE
		
		\STATE \textbf{Return} $\mathcal{G}$ 
	\end{algorithmic}
\end{algorithm}

\section{Experiments}\label{sec:chapEntropyBased_exp}
In this section, we evaluate our method experimentally on several datasets. We propose first an extensive analysis on simulated data with equal and different sampling rates, generated from basic causal structures; then, we performed an analysis on real-world datasets.
To assess the quality of causal inference, we used the F1-score regarding directed edges in the graph without considering self-loops.

In the following, we first describe the different settings of the methods we compare with and the datasets, and then, we describe the results and provide a complexity analysis.

\subsection{Methods and Their Use}
We compared our method \texttt{PCTMI}, available at \url{https://github.com/ckassaad/PCTMI} ({accessed on 30 June 2022}),%MDPI: Please add the access date (Format: Date Month Year). e.g., (accessed on 1 January 2020).
with several methods.
From the Granger family, we ran \texttt{MVGC} through the implementation available at \url{http://www.sussex.ac.uk/sackler/mvgc/} ({accessed on 30 June 2022}) and \texttt{TCDF} through the implementation available at \url{https://github.com/M-Nauta/TCDF} ({accessed on 30 June 2022}). For \texttt{TCDF}, some hyperparameters have to be defined: we use da kernel of size $4$, a dilation coefficient $4$, one hidden layer, a learning rate of $0.01$, and $5000$ epochs.
{From the score-based family, we retained \texttt{Dynotears}, which is available at \url{https://github.com/quantumblacklabs/causalnex} ({accessed on 30 June 2022}), the hyperparameters of which were set to their recommended values (the regularization constants $\lambda_W = \lambda_A = 0.05$).}
Among the noise-based approaches, we ran \texttt{TiMINo}, which is available at \url{http://web.math.ku.dk/~peters/code.html} ({accessed on 30 June 2022}). \texttt{TiMINo} uses a test based on the cross-correlation.% that can be derived from \citep[Thm 11.2.3.]{Brockwell_1986}.
%We also run the hybrid method \texttt{NBCB} for which we use a Gaussian Process with zero mean and squared exponential covariance function. The hyper-parameters are automatically chosen by marginal likelihood optimization.
From the constraint-based family, we ran \texttt{PCMCI} using the mutual information to measure the dependence, both provided in the implementation available at \url{https://github.com/jakobrunge/tigramite} ({accessed on 30 June 2022}). %We also use \texttt{oCSE}, which we implement. 

We compared \texttt{FCITMI} with \texttt{tsFCI}, provided at \url{https://sites.google.com/site/dorisentner/ publications/tsfci} ({accessed on 30 June 2022}), where independence or conditional independence is tested, respectively, with tests of zero correlation or zero partial correlation.

For all methods using mutual information, we used the k-nearest neighbor estimator~\citep{Frenzel_2007}, for which we fixed the number of nearest neighbor to $k=10$. Since the output of those measures are necessarily positive given a finite sample size and a finite numerical precision, we used a significance permutation test introduced in \cite{Runge_2018}.
For all methods, we used $\gamma_{\max} = 5$, and when performing a statistical test, we used a significance level of $0.05$.

\subsection{Datasets}
\label{sec:exp_dataset}
To illustrate the behavior of the causal inference algorithms, we relied on both artificial and real-world datasets.
\subsubsection{Simulated data}
\label{subsec:sim_data}

In the case of causal sufficiency, we simulated time series with equal and different sampling rates of size $1000$ generated from three different causal structures: fork, v structure, and diamond represented in Table~\ref{tab:structure}. In the case of non-causal sufficiency, we considered the 7ts2h structure represented in Table~~\ref{tab:structure_hidden}.

For each structure, we generated 10 datasets using the following generating process: for all $q$, $X^q_0 = 0$, and for all $t>0$,
%Each time series $X^{q}$ in these structures is generated from its parents according to 
%
{\small{
		\begin{equation*}
		X^{q}_t = a^{qq}_{t-1} X^{q}_{t-1} + \sum_{(p, \gamma) \atop X^{p}_{t-\gamma} \in Par(X^{q}_t)} a^{pq}_{t-\gamma} f(X^{p}_{t-\gamma}) + 0.1 \xi^q_t,
		\end{equation*}
}}
where $\gamma\geq 0$, $a^{jq}_t$ are random coefficients chosen uniformly in $ \mathcal{U}([-1; - 0.1 ] \cup [ 0.1; 1 ])$ for all $1\leq j \leq d$, $\xi^q_t \sim \mathcal{N}(0, \sqrt{15})$ and $f$ is a nonlinear function chosen at random uniformly among the absolute value, tanh, sine, and cosine.

\begin{table}[H]%[hbt]
	\caption[Structures of simulated data without hidden common causes.]{Structures of simulated data without hidden common causes. $A\rightarrow B$ means that $A$ causes $B$.}
	\label{tab:structure}
	\setlength{\tabcolsep}{10pt} % separator between columns
	\def\arraystretch{1.25} % vertical stretch factor
	%\centering
	\setlength{\tabcolsep}{9.7mm}{\begin{tabular}{ccc}
			%\hline
			\toprule
			\textbf{V-Structure} & \textbf{Fork} & \textbf{Diamond}
			\\ 
			\midrule
			\begin{tikzpicture}[{black, circle, draw, inner sep=0}]
			\tikzset{nodes={draw,rounded corners},minimum height=0.5cm,minimum width=0.5cm, font=\footnotesize}
			
			\node (X) at (0,0) {$ X^{1}$} ;
			\node (Y) at (2,0) {$ X^{2}$} ;
			\node (Z) at (1,-0.5) {$ X^{3}$};
			\draw[->,>=latex] (X) -- (Z);
			\draw[->,>=latex] (Y) -- (Z);
			\draw[->,>=latex] (X) to [out=0,in=45, looseness=2] (X);
			\draw[->,>=latex] (Y) to [out=0,in=45, looseness=2] (Y);
			\draw[->,>=latex] (Z) to [out=0,in=45, looseness=2] (Z);
			\end{tikzpicture} &
			\begin{tikzpicture}[{black, circle, draw, inner sep=0}]
			\tikzset{nodes={draw,rounded corners},minimum height=0.5cm,minimum width=0.5cm, font=\footnotesize}
			\node (X) at (1,0) {$ X^{1}$} ;
			\node (Y) at (0,-0.5) {$ X^{2}$} ;
			\node (Z) at (2,-0.5) {$ X^{3}$};
			\draw[->,>=latex] (X) -- (Y);
			\draw[->,>=latex] (X) -- (Z);
			\draw[->,>=latex] (X) to [out=0,in=45, looseness=2] (X);
			\draw[->,>=latex] (Y) to [out=0,in=45, looseness=2] (Y);
			\draw[->,>=latex] (Z) to [out=0,in=45, looseness=2] (Z);
			\end{tikzpicture}
			& 
			\begin{tikzpicture}[{black, circle, draw, inner sep=0}]
			\tikzset{nodes={draw,rounded corners},minimum height=0.5cm,minimum width=0.5cm, font=\footnotesize}
			\node (X) at (1,0) {$ X^{1}$} ;
			\node (Y) at (0,-0.5) {$ X^{2}$} ;
			\node (Z) at (2,-0.5) {$ X^{3}$};
			\node (W) at (1,-0.7) {$ X^{4}$};
			\draw[->,>=latex] (X) -- (Y);
			\draw[->,>=latex] (X) -- (Z);
			\draw[->,>=latex] (Y) -- (W);
			\draw[->,>=latex] (Z) -- (W);
			\draw[->,>=latex] (X) to [out=0,in=45, looseness=2] (X);
			\draw[->,>=latex] (Y) to [out=0,in=45, looseness=2] (Y);
			\draw[->,>=latex] (Z) to [out=0,in=45, looseness=2] (Z);
			\draw[->,>=latex] (W) to [out=0,in=45, looseness=2] (W);
			\end{tikzpicture}\\\bottomrule
			%\bottomrule
	\end{tabular}}
\end{table}
\vspace{-6pt}

\begin{table}[H]%[hbt]
	\caption[Structures of simulated data with hidden common causes.]{Structures of simulated data with hidden common causes. $A\rightarrow B$ means that A causes B and $A \rightlefta$ B represents the existence of a hidden common cause between A and B.}
	\label{tab:structure_hidden}
	\setlength{\tabcolsep}{10pt} % separator between columns
	\def\arraystretch{1.25} % vertical stretch factor
	%\centering
	\setlength{\tabcolsep}{55mm}{\begin{tabular}{c}
			%\hline
			\toprule
			\textbf{7ts2h} 
			\\ 
			\midrule
			\begin{tikzpicture}[{black, circle, draw, inner sep=0}]
			\tikzset{nodes={draw,rounded corners},minimum height=0.5cm,minimum width=0.5cm, font=\footnotesize}
			%		\tikzset{latent/.append style={fill=gray!30}}
			
			%		\node [latent] (T1) at (0.5,1) {$T1$};
			%		\node [latent] (T2) at (0.5,-2) {$T2$};
			\node (E) at (1.5,-0.5) {$ X^{1}$} ;
			\node (B) at (1,0.5) {$ X^{2}$} ;
			\node (F) at (2,0.5) {$ X^{3}$} ;
			\node (C) at (2.5,-0.5) {$ X^{4}$} ;
			\node (H) at (2,-1.5) {$ X^{5}$} ;
			\node (D) at (1,-1.5) {$ X^{6}$} ;
			\node (A) at (0.5,-0.5) {$ X^{7}$} ;
			%		\draw[->,>=latex] (T1) -- (A);
			%		\draw[->,>=latex] (T1) -- (B);
			\draw[<->,>=latex] (A) -- (B);
			\draw[->,>=latex] (B) -- (A);
			\draw[->,>=latex] (B) -- (E);
			\draw[->,>=latex] (F) -- (B);
			\draw[->,>=latex] (C) -- (F);
			\draw[->,>=latex] (C) -- (H);
			\draw[->,>=latex] (H) -- (D);
			\draw[->,>=latex] (D) -- (A);
			\draw[<->,>=latex] (E) -- (D);
			
			\draw[->,>=latex] (A) to [out=180,in=135, looseness=2] (A);
			\draw[->,>=latex] (B) to [out=0,in=45, looseness=2] (B);
			\draw[->,>=latex] (C) to [out=0,in=45, looseness=2] (C);
			\draw[->,>=latex] (D) to [out=0,in=45, looseness=2] (D);
			\draw[->,>=latex] (E) to [out=180,in=135, looseness=2] (E);
			\draw[->,>=latex] (F) to [out=0,in=45, looseness=2] (F);
			\draw[->,>=latex] (H) to [out=0,in=45, looseness=2] (H);
			%		\draw[->,>=latex] (T2) -- (D);
			%		\draw[->,>=latex] (T2) -- (E);
			\end{tikzpicture}\\\bottomrule
	\end{tabular}}
\end{table}

\subsubsection{Real Data}
First, we considered the realistic functional magnetic resonance imaging (fMRI) benchmark, which contains blood-oxygen-level dependent (BOLD) datasets \citep{Smith_2011} for 28 different underlying brain networks. It measures the neural activity of different regions of interest in the brain based on the change of blood flow. 
There are 50 regions in total, each with its own associated time series.
Since not all existing methods can handle 50 time series, datasets with more than 10 time series were excluded. 
In total, we were left with 26 datasets containing between 5 and 10 brain regions.

Second, we considered time series collected from an IT monitoring system with a one-minute sampling rate provided by EasyVista {(}\url{https://www.easyvista.com/fr/produits/servicenav}, {accessed on 30 June 2022}{)}. %MDPI: The footnote is not allowed, moved into the text, please conform. (Confirmed) 
In total, we considered $24$ datasets with $1000$ timestamps. Each dataset contains three time series: the \emph{{metric extraction}} (M.E), which represents the activity of the extraction of the metrics from the messages; the \emph{group history insertion} (G.H.I), which represents the activity of the insertion of the historical status in the database; and the \emph{collector monitoring information} (C.M.I), which represents the activity of the updates in a given database. 
Lags between time series are unknown, as well as the existence of self-causes.
According to the domain experts, all these datasets follow a fork structure such that \emph{metric extraction} is a common cause of the other two time series.%MDPI: Please confirm if the italic should be retained, and confirm this type in the full text (Confirmed)

\subsection{Numerical Results}
\subsubsection{Simulated Data}

We provide in Table~\ref{tab:sim} the performance of all methods on simulated data with an equal sampling rate.
\texttt{PCTMI} has better results than other methods for the v structure and the fork structure 
%and is outperformed only by \texttt{oCSE} for other structures. High performance of \texttt{oCSE} is related to strong assumptions used by this method namely that a cause relation between two different time series is necessary 1-order Markov. 
For the diamond structure, \texttt{PCTMI} and \texttt{PCMCI} have the same performance; however, \texttt{PCTMI} has a lower standard deviation.
In general, we can say, that constraint-based algorithms perform best.
\begin{table}[H]
	\caption{Results for simulated datasets with an equal sampling rate. We report the mean and the standard deviation of the F1-score. The best results are in bold.} \label{tab:sim}
	\centering
	\begin{adjustwidth}{-0.41\linewidth}{0cm}
		\newcolumntype{C}{>{\centering\arraybackslash}X}
		\begin{tabularx}{\linewidth}{CCCCCCCC}
			\toprule
			&\texttt{\textbf{PCTMI}} & 	\texttt{\textbf{PCMCI}} & \texttt{\textbf{TiMINo}} & \texttt{\textbf{VarLiNGAM}} & \texttt{\textbf{Dynotears}} & \texttt{\textbf{TCDF}} & \texttt{\textbf{MVGC}} \\
			\midrule
			V structure& $\textbf{0.78} \pm 0.18$ & $0.67 \pm 0.37$ & $0.65 \pm 0.37$ & $0.0 \pm 0.0$ & $0.07 \pm 0.20$ &$0.13 \pm 0.26$ &$0.37 \pm 0.26$ \\
			Fork& $\textbf{0.83} \pm 0.31$ & $0.78 \pm 0.17$ & $0.52 \pm 0.44$ & $0.0 \pm 0.0$ & $0.07 \pm 0.20$ &$0.26 \pm 0.32$ &$0.44 \pm 0.38$ \\
			Diamond & $\textbf{0.82} \pm 0.11$ & $\textbf{0.82} \pm 0.16$ & $0.60 \pm 0.25$ & $0.03 \pm 0.09$ & $0.23 \pm 0.24$ &$0.16 \pm 0.19$ &$0.68 \pm 0.26$ \\
			\bottomrule
		\end{tabularx} 
	\end{adjustwidth}
\end{table}
% oCSE V structure: $\textbf{0.90} \pm 0.16$, Fork: $0.80 \pm 0.12$, Diamond:$\textbf{0.88} \pm 0.09$
%NBCB V structure: $0.67 \pm 0.28$, Fork: $0.67 \pm 0.38$, Diamond:$0.60 \pm 0.25$

%\newcolumntype{C}{>{\centering\arraybackslash}X}
%\begin{tabularx}{\textwidth}{CCC}
%	\toprule
%	\textbf{Title 1}	& \textbf{Title 2}	& \textbf{Title 3}\\
%	\midrule
%	Entry 1		& Data			& Data\\
%	Entry 2		& Data			& Data\\
%	\bottomrule
%\end{tabularx}

We also assess the behavior of \texttt{PCTMI} when the time series have different sampling rates in Table~\ref{tab:sim_diff}. We present here results only for \texttt{PCTMI}, because
other methods are not applicable to time series with different sampling rates. 
As one can see, the performance obtained here is close to the ones obtained with equal sampling rates in the case of the v structure, but significantly lower in the case if the fork structure and the diamond. 
The degradation of the results is not really surprising as one has less data to rely on in a different sampling rate scenario.
By comparing the results of \texttt{PCTMI} on time series with different sampling rates with the results of other methods obtained when time series have an equal sampling rate, we can see that \texttt{PCTMI} still performs better than most methods.

\begin{table}[H]
	%\begin{wraptable}{r}{0.32\textwidth}
	\caption{Results for simulated datasets with different sampling rates. We report the mean and the standard deviation of the F1-score.} \label{tab:sim_diff}
	\centering
	\newcolumntype{C}{>{\centering\arraybackslash}X}
	\begin{tabularx}{\textwidth}{CC}
		\toprule
		&\texttt{\textbf{PCTMI}} \\
		\midrule
		V structure& $0.80 \pm 0.31$\\
		Fork& $0.56 \pm 0.30$\\
		Diamond & $0.66 \pm 0.24$\\
		\bottomrule
	\end{tabularx} 
	%\end{wraptable}
\end{table}

In the case of causal non-sufficiency, our proposed algorithm \texttt{FCITMI} outperforms \texttt{tsFCI}, as shown in Table~\ref{tab:sim_hidd}.

\begin{table}[H]
	%\begin{wraptable}{r}{0.32\textwidth}
	\caption{Results for simulated datasets with an equal sampling rate and with hidden common causes. We report the mean and the standard deviation of the F1-score.} \label{tab:sim_hidd}
	%\centering
	\newcolumntype{C}{>{\centering\arraybackslash}X}
	\begin{tabularx}{\textwidth}{CCC}
		\toprule
		&\texttt{\textbf{FCITMI}} & \texttt{\textbf{tsFCI}} \\
		\midrule
		7ts2h& $0.44 \pm 0.11$ & $0.37 \pm 0.09$\\
		\bottomrule
	\end{tabularx} 
	%\end{wraptable}
\end{table}

\subsubsection{Real Data}
We provide in Table \ref{tab:FMRI} the results on the {fMRI} benchmark and the IT benchmark for all the methods.
In the case of {fMRI}, \texttt{VarLiNGAM} performs best, followed by \texttt{Dynotears} and, then, by \texttt{PCTMI} and \texttt{TiMINo}. The success of linear methods (\texttt{VarLiNGAM} and \texttt{Dynotears}) on this benchmark might suggest that causal relations in {fMRI} are linear, especially as this result was not replicated on other datasets considered in this paper.
In the case of the IT datasets, \texttt{TiMINo} performs best, followed by \texttt{PCTMI}; however, while investigating the results, we noticed that for $6/10$ datasets, \texttt{TiMINo} returns a fully connected summary graph with arcs oriented from both sides.
In both benchmarks, \texttt{PCTMI} clearly outperforms other constraint-based~methods.

\begin{table}[H]
	\caption{Results for real datasets. We report the mean and the standard deviation of the F1-score. The best results are in bold.} \label{tab:FMRI}
	\centering
	\begin{adjustwidth}{-0.41\linewidth}{0cm}
		\newcolumntype{C}{>{\centering\arraybackslash}X}
		\begin{tabularx}{\linewidth}{CCCCCCCC}
			\toprule 
			& \texttt{\textbf{PCTMI}} & 	\textbf{\texttt{PCMCI}} & \texttt{\textbf{TiMINo}}& \texttt{\textbf{VarLiNGAM}} & \texttt{\textbf{Dynotears}} & \texttt{\textbf{TCDF}} & \texttt{\textbf{MVGC}}\\
			\midrule
			{fMRI} & $0.32 \pm 0.17$ & $0.22 \pm 0.18$ & $0.32 \pm 0.11$ & $\textbf{0.49} \pm 0.28$ & $0.34 \pm 0.13$ &$0.07 \pm 0.13$ & $0.24 \pm 0.18$\\
			IT& $0.40 \pm 0.32$ & $0.25 \pm 0.31$ & $\textbf{0.62} \pm 0.14$ & $0.36 \pm 0.19$ & $0.0 \pm 0.0$ & $0.0 \pm 0.0$ & $0.38 \pm 0.17$\\
			%		IT chain & $0.0 \pm 0.0$ & $0.0 \pm 0.0$ & $0.0 \pm 0.0$ & $0.0 \pm 0.0$ & $0.0 \pm 0.0$ & $0.0 \pm 0.0$ & $0.0 \pm 0.0$ & $0.0 \pm 0.0$\\
			\bottomrule
		\end{tabularx} 
	\end{adjustwidth}
\end{table}
% oCSE FMRI:$0.16 \pm 0.20$ , IT: $0.64 \pm 0.26$ 
% NBCB FMRI:$0.4 \pm 0.21$ , IT: $0.4 \pm 0.35$ 

In the case of the IT datasets, we also provide in {Table~\ref{tab:res_it_details}, for each dataset, the inferred summary causal graphs by methods that have an F1-score $>0$ in Table~\ref{tab:FMRI}. Here, we can see that in $5$ out of $10$ datasets, \texttt{PCTMI} was able to infer that M.E is a cause of G.H.I. On the other hand, \texttt{PCTMI} was able to infer that M.E is a cause of C.M.I. in only one dataset. Interestingly, in $9$ out $10$ datasets, \texttt{PCTMI} did not infer any false positive. 
	As one can expect from Table~\ref{tab:FMRI}, \texttt{PCMCI} suffers on all datasets; however, one can notice that as \texttt{PCTMI}, \texttt{PCMCI} has a tendency to yield sparse graphs.
	\texttt{TiMINo}, which has the best F1-score in Table~\ref{tab:FMRI}, seems to have a very low precision. In $7$ out $10$ datasets, \texttt{TiMINo} returned a complete bidirected graph. In the other three datasets, \texttt{TiMINo} detects all true causal relations, in addition to one false positive.
	\texttt{VarLiNGAM} has an even lower precision compared to \texttt{TiMINo} and a lower recall. Finally, \texttt{MVGC} infers most of the time a complete bidirected graph.
}

\begin{table}[H]%[hbt]
	\caption{{Summary causal graphs inferred by different methods using 10 different monitoring IT datasets. $A\rightarrow B$ means that $A$ causes $B$ and $A \rightleftarrows B$ means that $A$ causes $B$ and $B$ causes $A$.}}
	\label{tab:res_it_details}
	\begin{adjustwidth}{-0.41\linewidth}{0cm}
		\newcolumntype{C}{>{\centering\arraybackslash}X}
		\begin{tabularx}{\linewidth}{CCCCCCCC}
			\toprule 
			%\hline
			& \texttt{PCTMI} & \texttt{PCMCI} & \texttt{TiMINo} & \texttt{VarLiNGAM} & \texttt{MVGC}
			\\ 
			\midrule
			Dataset 1 &
			\begin{tikzpicture}[{black, circle, draw, inner sep=0}]
			\tikzset{nodes=minimum height=0.5cm,minimum width=0.5cm, font=\footnotesize}
			\node (X) at (1,0) {M.E} ;
			\node (Y) at (0,-0.5) {G.H.I} ;
			\node (Z) at (2,-0.5) {C.M.I};
			\draw[->,>=latex] (X) -- (Y);
			%				\draw[->,>=latex] (X) -- (Z);
			\end{tikzpicture} 
			&
			\begin{tikzpicture}[{black, circle, draw, inner sep=0}]
			\tikzset{nodes=minimum height=0.5cm,minimum width=0.5cm, font=\footnotesize}
			\node (X) at (1,0) {M.E} ;
			\node (Y) at (0,-0.5) {G.H.I} ;
			\node (Z) at (2,-0.5) {C.M.I};
			\draw[->,>=latex] (X) to [out=180,in=45, looseness=1] (Y);
			\draw[->,>=latex] (Y) to [out=0,in=-135, looseness=1] (X);
			\end{tikzpicture}
			& 
			\begin{tikzpicture}[{black, circle, draw, inner sep=0}]
			\tikzset{nodes=minimum height=0.5cm,minimum width=0.5cm, font=\footnotesize}
			\node (X) at (1,0) {M.E} ;
			\node (Y) at (0,-0.5) {G.H.I} ;
			\node (Z) at (2,-0.5) {C.M.I};
			\draw[->,>=latex] (X) to [out=180,in=45, looseness=1] (Y);
			\draw[->,>=latex] (Y) to [out=0,in=-135, looseness=1] (X);
			\draw[->,>=latex] (X) to [out=0,in=135, looseness=1] (Z);
			\draw[->,>=latex] (Z) to [out=180,in=-45, looseness=1] (X);
			\draw[->,>=latex] (Z) to [out=-145,in=-35, looseness=1] (Y);
			\draw[->,>=latex] (Y) to [out=-10,in=190, looseness=1] (Z);
			\end{tikzpicture}
			&
			\begin{tikzpicture}[{black, circle, draw, inner sep=0}]
			\tikzset{nodes=minimum height=0.5cm,minimum width=0.5cm, font=\footnotesize}
			\node (X) at (1,0) {M.E} ;
			\node (Y) at (0,-0.5) {G.H.I} ;
			\node (Z) at (2,-0.5) {C.M.I};
			\draw[->,>=latex] (Y) -- (X);
			\draw[->,>=latex] (Z) -- (X);
			\draw[->,>=latex] (Y) -- (Z);
			\end{tikzpicture}
			&
			\begin{tikzpicture}[{black, circle, draw, inner sep=0}]
			\tikzset{nodes=minimum height=0.5cm,minimum width=0.5cm, font=\footnotesize}
			\node (X) at (1,0) {M.E} ;
			\node (Y) at (0,-0.5) {G.H.I} ;
			\node (Z) at (2,-0.5) {C.M.I};
			\draw[->,>=latex] (X) -- (Y);
			\draw[->,>=latex] (Y) -- (Z);
			\end{tikzpicture}
			\\\midrule
			Dataset 2 &
			\begin{tikzpicture}[{black, circle, draw, inner sep=0}]
			\tikzset{nodes=minimum height=0.5cm,minimum width=0.5cm, font=\footnotesize}
			\node (X) at (1,0) {M.E} ;
			\node (Y) at (0,-0.5) {G.H.I} ;
			\node (Z) at (2,-0.5) {C.M.I};
			\draw[->,>=latex] (X) -- (Y);
			%				\draw[->,>=latex] (X) -- (Z);
			\end{tikzpicture} 
			&
			\begin{tikzpicture}[{black, circle, draw, inner sep=0}]
			\tikzset{nodes=minimum height=0.5cm,minimum width=0.5cm, font=\footnotesize}
			\node (X) at (1,0) {M.E} ;
			\node (Y) at (0,-0.5) {G.H.I} ;
			\node (Z) at (2,-0.5) {C.M.I};
			\end{tikzpicture}
			& 
			\begin{tikzpicture}[{black, circle, draw, inner sep=0}]
			\tikzset{nodes=minimum height=0.5cm,minimum width=0.5cm, font=\footnotesize}
			\node (X) at (1,0) {M.E} ;
			\node (Y) at (0,-0.5) {G.H.I} ;
			\node (Z) at (2,-0.5) {C.M.I};
			\draw[->,>=latex] (X) to [out=180,in=45, looseness=1] (Y);
			\draw[->,>=latex] (Y) to [out=0,in=-135, looseness=1] (X);
			\draw[->,>=latex] (X) to [out=0,in=135, looseness=1] (Z);
			\draw[->,>=latex] (Z) to [out=180,in=-45, looseness=1] (X);
			\draw[->,>=latex] (Z) to [out=-145,in=-35, looseness=1] (Y);
			\draw[->,>=latex] (Y) to [out=-10,in=190, looseness=1] (Z);
			\end{tikzpicture}
			&
			\begin{tikzpicture}[{black, circle, draw, inner sep=0}]
			\tikzset{nodes=minimum height=0.5cm,minimum width=0.5cm, font=\footnotesize}
			\node (X) at (1,0) {M.E} ;
			\node (Y) at (0,-0.5) {G.H.I} ;
			\node (Z) at (2,-0.5) {C.M.I};
			\draw[->,>=latex] (X) to [out=180,in=45, looseness=1] (Y);
			\draw[->,>=latex] (Y) to [out=0,in=-135, looseness=1] (X);
			\draw[->,>=latex] (Z) to [out=-145,in=-35, looseness=1] (Y);
			\draw[->,>=latex] (Y) to [out=-10,in=190, looseness=1] (Z);
			\end{tikzpicture}
			&
			\begin{tikzpicture}[{black, circle, draw, inner sep=0}]
			\tikzset{nodes=minimum height=0.5cm,minimum width=0.5cm, font=\footnotesize}
			\node (X) at (1,0) {M.E} ;
			\node (Y) at (0,-0.5) {G.H.I} ;
			\node (Z) at (2,-0.5) {C.M.I};
			\draw[->,>=latex] (X) -- (Y);
			\draw[->,>=latex] (Z) to [out=-145,in=-35, looseness=1] (Y);
			\draw[->,>=latex] (Y) to [out=-10,in=190, looseness=1] (Z);
			\end{tikzpicture}
			\\\midrule
			Dataset 3 &
			\begin{tikzpicture}[{black, circle, draw, inner sep=0}]
			\tikzset{nodes=minimum height=0.5cm,minimum width=0.5cm, font=\footnotesize}
			\node (X) at (1,0) {M.E} ;
			\node (Y) at (0,-0.5) {G.H.I} ;
			\node (Z) at (2,-0.5) {C.M.I};
			\draw[->,>=latex] (X) -- (Y);
			%				\draw[->,>=latex] (X) -- (Z);
			\end{tikzpicture} 
			&
			\begin{tikzpicture}[{black, circle, draw, inner sep=0}]
			\tikzset{nodes=minimum height=0.5cm,minimum width=0.5cm, font=\footnotesize}
			\node (X) at (1,0) {M.E} ;
			\node (Y) at (0,-0.5) {G.H.I} ;
			\node (Z) at (2,-0.5) {C.M.I};
			\draw[->,>=latex] (Y) -- (Z);
			\end{tikzpicture}
			& 
			\begin{tikzpicture}[{black, circle, draw, inner sep=0}]
			\tikzset{nodes=minimum height=0.5cm,minimum width=0.5cm, font=\footnotesize}
			\node (X) at (1,0) {M.E} ;
			\node (Y) at (0,-0.5) {G.H.I} ;
			\node (Z) at (2,-0.5) {C.M.I};
			\draw[->,>=latex] (X) to [out=180,in=45, looseness=1] (Y);
			\draw[->,>=latex] (Y) to [out=0,in=-135, looseness=1] (X);
			\draw[->,>=latex] (X) to [out=0,in=135, looseness=1] (Z);
			\draw[->,>=latex] (Z) to [out=180,in=-45, looseness=1] (X);
			\draw[->,>=latex] (Z) to [out=-145,in=-35, looseness=1] (Y);
			\draw[->,>=latex] (Y) to [out=-10,in=190, looseness=1] (Z);
			\end{tikzpicture}
			&
			\begin{tikzpicture}[{black, circle, draw, inner sep=0}]
			\tikzset{nodes=minimum height=0.5cm,minimum width=0.5cm, font=\footnotesize}
			\node (X) at (1,0) {M.E} ;
			\node (Y) at (0,-0.5) {G.H.I} ;
			\node (Z) at (2,-0.5) {C.M.I};
			\draw[->,>=latex] (X) -- (Y);
			\draw[->,>=latex] (Z) -- (Y);
			\end{tikzpicture}
			&
			\begin{tikzpicture}[{black, circle, draw, inner sep=0}]
			\tikzset{nodes=minimum height=0.5cm,minimum width=0.5cm, font=\footnotesize}
			\node (X) at (1,0) {M.E} ;
			\node (Y) at (0,-0.5) {G.H.I} ;
			\node (Z) at (2,-0.5) {C.M.I};
			\draw[->,>=latex] (X) to [out=180,in=45, looseness=1] (Y);
			\draw[->,>=latex] (Y) to [out=0,in=-135, looseness=1] (X);
			\draw[->,>=latex] (X) to [out=0,in=135, looseness=1] (Z);
			\draw[->,>=latex] (Z) to [out=180,in=-45, looseness=1] (X);
			\draw[->,>=latex] (Z) to [out=-145,in=-35, looseness=1] (Y);
			\draw[->,>=latex] (Y) to [out=-10,in=190, looseness=1] (Z);
			\end{tikzpicture}
			\\\midrule
			Dataset 4 &
			\begin{tikzpicture}[{black, circle, draw, inner sep=0}]
			\tikzset{nodes=minimum height=0.5cm,minimum width=0.5cm, font=\footnotesize}
			\node (X) at (1,0) {M.E} ;
			\node (Y) at (0,-0.5) {G.H.I} ;
			\node (Z) at (2,-0.5) {C.M.I};
			\draw[->,>=latex] (X) -- (Y);
			%				\draw[->,>=latex] (X) -- (Z);
			\end{tikzpicture} 
			&
			\begin{tikzpicture}[{black, circle, draw, inner sep=0}]
			\tikzset{nodes=minimum height=0.5cm,minimum width=0.5cm, font=\footnotesize}
			\node (X) at (1,0) {M.E} ;
			\node (Y) at (0,-0.5) {G.H.I} ;
			\node (Z) at (2,-0.5) {C.M.I};
			\draw[->,>=latex] (Y) -- (Z);
			\end{tikzpicture}
			& 
			\begin{tikzpicture}[{black, circle, draw, inner sep=0}]
			\tikzset{nodes=minimum height=0.5cm,minimum width=0.5cm, font=\footnotesize}
			\node (X) at (1,0) {M.E} ;
			\node (Y) at (0,-0.5) {G.H.I} ;
			\node (Z) at (2,-0.5) {C.M.I};
			\draw[->,>=latex] (X) to [out=180,in=45, looseness=1] (Y);
			\draw[->,>=latex] (Y) to [out=0,in=-135, looseness=1] (X);
			\draw[->,>=latex] (X) to [out=0,in=135, looseness=1] (Z);
			\draw[->,>=latex] (Z) to [out=180,in=-45, looseness=1] (X);
			\draw[->,>=latex] (Z) to [out=-145,in=-35, looseness=1] (Y);
			\draw[->,>=latex] (Y) to [out=-10,in=190, looseness=1] (Z);
			\end{tikzpicture}
			&
			\begin{tikzpicture}[{black, circle, draw, inner sep=0}]
			\tikzset{nodes=minimum height=0.5cm,minimum width=0.5cm, font=\footnotesize}
			\node (X) at (1,0) {M.E} ;
			\node (Y) at (0,-0.5) {G.H.I} ;
			\node (Z) at (2,-0.5) {C.M.I};
			\draw[->,>=latex] (X) to [out=180,in=45, looseness=1] (Y);
			\draw[->,>=latex] (Y) to [out=0,in=-135, looseness=1] (X);
			\draw[->,>=latex] (X) to [out=0,in=135, looseness=1] (Z);
			\draw[->,>=latex] (Z) to [out=180,in=-45, looseness=1] (X);
			\draw[->,>=latex] (Y) to [out=-10,in=190, looseness=1] (Z);
			\end{tikzpicture}
			&
			\begin{tikzpicture}[{black, circle, draw, inner sep=0}]
			\tikzset{nodes=minimum height=0.5cm,minimum width=0.5cm, font=\footnotesize}
			\node (X) at (1,0) {M.E} ;
			\node (Y) at (0,-0.5) {G.H.I} ;
			\node (Z) at (2,-0.5) {C.M.I};
			\draw[->,>=latex] (X) to [out=180,in=45, looseness=1] (Y);
			\draw[->,>=latex] (Y) to [out=0,in=-135, looseness=1] (X);
			\draw[->,>=latex] (X) to [out=0,in=135, looseness=1] (Z);
			\draw[->,>=latex] (Z) to [out=180,in=-45, looseness=1] (X);
			\draw[->,>=latex] (Z) to [out=-145,in=-35, looseness=1] (Y);
			\draw[->,>=latex] (Y) to [out=-10,in=190, looseness=1] (Z);
			\end{tikzpicture}
			\\\midrule
			Dataset 5 &
			\begin{tikzpicture}[{black, circle, draw, inner sep=0}]
			\tikzset{nodes=minimum height=0.5cm,minimum width=0.5cm, font=\footnotesize}
			\node (X) at (1,0) {M.E} ;
			\node (Y) at (0,-0.5) {G.H.I} ;
			\node (Z) at (2,-0.5) {C.M.I};
			%				\draw[->,>=latex] (X) -- (Y);
			%				\draw[->,>=latex] (X) -- (Z);
			\end{tikzpicture} 
			&
			\begin{tikzpicture}[{black, circle, draw, inner sep=0}]
			\tikzset{nodes=minimum height=0.5cm,minimum width=0.5cm, font=\footnotesize}
			\node (X) at (1,0) {M.E} ;
			\node (Y) at (0,-0.5) {G.H.I} ;
			\node (Z) at (2,-0.5) {C.M.I};
			%				\draw[->,>=latex] (X) -- (Y);
			\draw[->,>=latex] (X) -- (Z);
			\end{tikzpicture}
			& 
			\begin{tikzpicture}[{black, circle, draw, inner sep=0}]
			\tikzset{nodes=minimum height=0.5cm,minimum width=0.5cm, font=\footnotesize}
			\node (X) at (1,0) {M.E} ;
			\node (Y) at (0,-0.5) {G.H.I} ;
			\node (Z) at (2,-0.5) {C.M.I};
			\draw[->,>=latex] (X) to [out=180,in=45, looseness=1] (Y);
			\draw[->,>=latex] (Y) to [out=0,in=-135, looseness=1] (X);
			\draw[->,>=latex] (X) -- (Z);
			\end{tikzpicture}
			&
			\begin{tikzpicture}[{black, circle, draw, inner sep=0}]
			\tikzset{nodes=minimum height=0.5cm,minimum width=0.5cm, font=\footnotesize}
			\node (X) at (1,0) {M.E} ;
			\node (Y) at (0,-0.5) {G.H.I} ;
			\node (Z) at (2,-0.5) {C.M.I};
			\draw[->,>=latex] (X) to [out=180,in=45, looseness=1] (Y);
			\draw[->,>=latex] (Y) to [out=0,in=-135, looseness=1] (X);
			\end{tikzpicture}
			&
			\begin{tikzpicture}[{black, circle, draw, inner sep=0}]
			\tikzset{nodes=minimum height=0.5cm,minimum width=0.5cm, font=\footnotesize}
			\node (X) at (1,0) {M.E} ;
			\node (Y) at (0,-0.5) {G.H.I} ;
			\node (Z) at (2,-0.5) {C.M.I};
			\draw[->,>=latex] (X) to [out=180,in=45, looseness=1] (Y);
			\draw[->,>=latex] (Y) to [out=0,in=-135, looseness=1] (X);
			\draw[->,>=latex] (X) to [out=0,in=135, looseness=1] (Z);
			\draw[->,>=latex] (Z) to [out=180,in=-45, looseness=1] (X);
			\draw[->,>=latex] (Z) to [out=-145,in=-35, looseness=1] (Y);
			\end{tikzpicture}
			\\ \midrule
			Dataset 6 &
			\begin{tikzpicture}[{black, circle, draw, inner sep=0}]
			\tikzset{nodes=minimum height=0.5cm,minimum width=0.5cm, font=\footnotesize}
			\node (X) at (1,0) {M.E} ;
			\node (Y) at (0,-0.5) {G.H.I} ;
			\node (Z) at (2,-0.5) {C.M.I};
			\draw[->,>=latex] (X) -- (Y);
			%				\draw[->,>=latex] (X) -- (Z);
			\end{tikzpicture} 
			&
			\begin{tikzpicture}[{black, circle, draw, inner sep=0}]
			\tikzset{nodes=minimum height=0.5cm,minimum width=0.5cm, font=\footnotesize}
			\node (X) at (1,0) {M.E} ;
			\node (Y) at (0,-0.5) {G.H.I} ;
			\node (Z) at (2,-0.5) {C.M.I};
			\draw[->,>=latex] (X) to [out=0,in=135, looseness=1] (Z);
			\draw[->,>=latex] (Z) to [out=180,in=-45, looseness=1] (X);
			\draw[->,>=latex] (Z) to [out=-145,in=-35, looseness=1] (Y);
			\draw[->,>=latex] (Y) to [out=-10,in=190, looseness=1] (Z);
			\end{tikzpicture}
			& 
			\begin{tikzpicture}[{black, circle, draw, inner sep=0}]
			\tikzset{nodes=minimum height=0.5cm,minimum width=0.5cm, font=\footnotesize}
			\node (X) at (1,0) {M.E} ;
			\node (Y) at (0,-0.5) {G.H.I} ;
			\node (Z) at (2,-0.5) {C.M.I};
			\draw[->,>=latex] (X) to [out=180,in=45, looseness=1] (Y);
			\draw[->,>=latex] (Y) to [out=0,in=-135, looseness=1] (X);
			\draw[->,>=latex] (X) to [out=0,in=135, looseness=1] (Z);
			\draw[->,>=latex] (Z) to [out=180,in=-45, looseness=1] (X);
			\draw[->,>=latex] (Z) to [out=-145,in=-35, looseness=1] (Y);
			\draw[->,>=latex] (Y) to [out=-10,in=190, looseness=1] (Z);
			\end{tikzpicture}
			&
			\begin{tikzpicture}[{black, circle, draw, inner sep=0}]
			\tikzset{nodes=minimum height=0.5cm,minimum width=0.5cm, font=\footnotesize}
			\node (X) at (1,0) {M.E} ;
			\node (Y) at (0,-0.5) {G.H.I} ;
			\node (Z) at (2,-0.5) {C.M.I};
			\draw[->,>=latex] (X) to [out=180,in=45, looseness=1] (Y);
			\draw[->,>=latex] (Y) to [out=0,in=-135, looseness=1] (X);
			\draw[->,>=latex] (X) to [out=0,in=135, looseness=1] (Z);
			\draw[->,>=latex] (Z) to [out=180,in=-45, looseness=1] (X);
			\draw[->,>=latex] (Z) to [out=-145,in=-35, looseness=1] (Y);
			\draw[->,>=latex] (Y) to [out=-10,in=190, looseness=1] (Z);
			\end{tikzpicture}
			&
			\begin{tikzpicture}[{black, circle, draw, inner sep=0}]
			\tikzset{nodes=minimum height=0.5cm,minimum width=0.5cm, font=\footnotesize}
			\node (X) at (1,0) {M.E} ;
			\node (Y) at (0,-0.5) {G.H.I} ;
			\node (Z) at (2,-0.5) {C.M.I};
			\draw[->,>=latex] (X) to [out=180,in=45, looseness=1] (Y);
			\draw[->,>=latex] (Y) to [out=0,in=-135, looseness=1] (X);
			\draw[->,>=latex] (X) to [out=0,in=135, looseness=1] (Z);
			\draw[->,>=latex] (Z) to [out=180,in=-45, looseness=1] (X);
			\draw[->,>=latex] (Z) to [out=-145,in=-35, looseness=1] (Y);
			\draw[->,>=latex] (Y) to [out=-10,in=190, looseness=1] (Z);
			\end{tikzpicture}
			\\
			\midrule
			Dataset 7 &
			\begin{tikzpicture}[{black, circle, draw, inner sep=0}]
			\tikzset{nodes=minimum height=0.5cm,minimum width=0.5cm, font=\footnotesize}
			\node (X) at (1,0) {M.E} ;
			\node (Y) at (0,-0.5) {G.H.I} ;
			\node (Z) at (2,-0.5) {C.M.I};
			%				\draw[->,>=latex] (X) -- (Y);
			\draw[->,>=latex] (X) -- (Z);
			\end{tikzpicture} 
			&
			\begin{tikzpicture}[{black, circle, draw, inner sep=0}]
			\tikzset{nodes=minimum height=0.5cm,minimum width=0.5cm, font=\footnotesize}
			\node (X) at (1,0) {M.E} ;
			\node (Y) at (0,-0.5) {G.H.I} ;
			\node (Z) at (2,-0.5) {C.M.I};
			\draw[->,>=latex] (Z) -- (X);
			\end{tikzpicture}
			& 
			\begin{tikzpicture}[{black, circle, draw, inner sep=0}]
			\tikzset{nodes=minimum height=0.5cm,minimum width=0.5cm, font=\footnotesize}
			\node (X) at (1,0) {M.E} ;
			\node (Y) at (0,-0.5) {G.H.I} ;
			\node (Z) at (2,-0.5) {C.M.I};
			\draw[->,>=latex] (X) to [out=180,in=45, looseness=1] (Y);
			\draw[->,>=latex] (Y) to [out=0,in=-135, looseness=1] (X);
			\draw[->,>=latex] (X) -- (Z);
			\end{tikzpicture}
			&
			\begin{tikzpicture}[{black, circle, draw, inner sep=0}]
			\tikzset{nodes=minimum height=0.5cm,minimum width=0.5cm, font=\footnotesize}
			\node (X) at (1,0) {M.E} ;
			\node (Y) at (0,-0.5) {G.H.I} ;
			\node (Z) at (2,-0.5) {C.M.I};
			\draw[->,>=latex] (X) to [out=180,in=45, looseness=1] (Y);
			\draw[->,>=latex] (Y) to [out=0,in=-135, looseness=1] (X);
			\draw[->,>=latex] (Z) -- (X);
			\end{tikzpicture}
			&
			\begin{tikzpicture}[{black, circle, draw, inner sep=0}]
			\tikzset{nodes=minimum height=0.5cm,minimum width=0.5cm, font=\footnotesize}
			\node (X) at (1,0) {M.E} ;
			\node (Y) at (0,-0.5) {G.H.I} ;
			\node (Z) at (2,-0.5) {C.M.I};
			\draw[->,>=latex] (X) to [out=180,in=45, looseness=1] (Y);
			\draw[->,>=latex] (Y) to [out=0,in=-135, looseness=1] (X);
			\draw[->,>=latex] (X) to [out=0,in=135, looseness=1] (Z);
			\draw[->,>=latex] (Z) to [out=180,in=-45, looseness=1] (X);
			\draw[->,>=latex] (Z) to [out=-145,in=-35, looseness=1] (Y);
			\draw[->,>=latex] (Y) to [out=-10,in=190, looseness=1] (Z);
			\end{tikzpicture}
			\\\midrule
			Dataset 8 &
			\begin{tikzpicture}[{black, circle, draw, inner sep=0}]
			\tikzset{nodes=minimum height=0.5cm,minimum width=0.5cm, font=\footnotesize}
			\node (X) at (1,0) {M.E} ;
			\node (Y) at (0,-0.5) {G.H.I} ;
			\node (Z) at (2,-0.5) {C.M.I};
			%				\draw[->,>=latex] (X) -- (Y);
			%				\draw[->,>=latex] (X) -- (Z);
			\end{tikzpicture} 
			&
			\begin{tikzpicture}[{black, circle, draw, inner sep=0}]
			\tikzset{nodes=minimum height=0.5cm,minimum width=0.5cm, font=\footnotesize}
			\node (X) at (1,0) {M.E} ;
			\node (Y) at (0,-0.5) {G.H.I} ;
			\node (Z) at (2,-0.5) {C.M.I};
			\draw[->,>=latex] (X) -- (Z);
			\end{tikzpicture}
			& 
			\begin{tikzpicture}[{black, circle, draw, inner sep=0}]
			\tikzset{nodes=minimum height=0.5cm,minimum width=0.5cm, font=\footnotesize}
			\node (X) at (1,0) {M.E} ;
			\node (Y) at (0,-0.5) {G.H.I} ;
			\node (Z) at (2,-0.5) {C.M.I};
			\draw[->,>=latex] (X) to [out=180,in=45, looseness=1] (Y);
			\draw[->,>=latex] (Y) to [out=0,in=-135, looseness=1] (X);
			\draw[->,>=latex] (X) -- (Z);
			\end{tikzpicture}
			&
			\begin{tikzpicture}[{black, circle, draw, inner sep=0}]
			\tikzset{nodes=minimum height=0.5cm,minimum width=0.5cm, font=\footnotesize}
			\node (X) at (1,0) {M.E} ;
			\node (Y) at (0,-0.5) {G.H.I} ;
			\node (Z) at (2,-0.5) {C.M.I};
			\draw[->,>=latex] (Y) -- (X);
			\end{tikzpicture}
			&
			\begin{tikzpicture}[{black, circle, draw, inner sep=0}]
			\tikzset{nodes=minimum height=0.5cm,minimum width=0.5cm, font=\footnotesize}
			\node (X) at (1,0) {M.E} ;
			\node (Y) at (0,-0.5) {G.H.I} ;
			\node (Z) at (2,-0.5) {C.M.I};
			\draw[->,>=latex] (X) to [out=180,in=45, looseness=1] (Y);
			\draw[->,>=latex] (Y) to [out=0,in=-135, looseness=1] (X);
			\draw[->,>=latex] (X) to [out=0,in=135, looseness=1] (Z);
			\draw[->,>=latex] (Z) to [out=180,in=-45, looseness=1] (X);
			\draw[->,>=latex] (Z) to [out=-145,in=-35, looseness=1] (Y);
			\draw[->,>=latex] (Y) to [out=-10,in=190, looseness=1] (Z);
			\end{tikzpicture}
			\\\midrule
			Dataset 9 &
			\begin{tikzpicture}[{black, circle, draw, inner sep=0}]
			\tikzset{nodes=minimum height=0.5cm,minimum width=0.5cm, font=\footnotesize}
			\node (X) at (1,0) {M.E} ;
			\node (Y) at (0,-0.5) {G.H.I} ;
			\node (Z) at (2,-0.5) {C.M.I};
			\draw[->,>=latex] (Y) -- (Z);
			%				\draw[->,>=latex] (X) -- (Z);
			\end{tikzpicture} 
			&
			\begin{tikzpicture}[{black, circle, draw, inner sep=0}]
			\tikzset{nodes=minimum height=0.5cm,minimum width=0.5cm, font=\footnotesize}
			\node (X) at (1,0) {M.E} ;
			\node (Y) at (0,-0.5) {G.H.I} ;
			\node (Z) at (2,-0.5) {C.M.I};
			\draw[->,>=latex] (X) -- (Y);
			\draw[->,>=latex] (Z) -- (X);
			\end{tikzpicture}
			& 
			\begin{tikzpicture}[{black, circle, draw, inner sep=0}]
			\tikzset{nodes=minimum height=0.5cm,minimum width=0.5cm, font=\footnotesize}
			\node (X) at (1,0) {M.E} ;
			\node (Y) at (0,-0.5) {G.H.I} ;
			\node (Z) at (2,-0.5) {C.M.I};
			\draw[->,>=latex] (X) to [out=180,in=45, looseness=1] (Y);
			\draw[->,>=latex] (Y) to [out=0,in=-135, looseness=1] (X);
			\draw[->,>=latex] (X) to [out=0,in=135, looseness=1] (Z);
			\draw[->,>=latex] (Z) to [out=180,in=-45, looseness=1] (X);
			\draw[->,>=latex] (Z) to [out=-145,in=-35, looseness=1] (Y);
			\draw[->,>=latex] (Y) to [out=-10,in=190, looseness=1] (Z);
			\end{tikzpicture}
			&
			\begin{tikzpicture}[{black, circle, draw, inner sep=0}]
			\tikzset{nodes=minimum height=0.5cm,minimum width=0.5cm, font=\footnotesize}
			\node (X) at (1,0) {M.E} ;
			\node (Y) at (0,-0.5) {G.H.I} ;
			\node (Z) at (2,-0.5) {C.M.I};
			\draw[->,>=latex] (X) to [out=180,in=45, looseness=1] (Y);
			\draw[->,>=latex] (Y) to [out=0,in=-135, looseness=1] (X);
			\draw[->,>=latex] (Z) -- (X);
			\end{tikzpicture}
			&
			\begin{tikzpicture}[{black, circle, draw, inner sep=0}]
			\tikzset{nodes=minimum height=0.5cm,minimum width=0.5cm, font=\footnotesize}
			\node (X) at (1,0) {M.E} ;
			\node (Y) at (0,-0.5) {G.H.I} ;
			\node (Z) at (2,-0.5) {C.M.I};
			\draw[->,>=latex] (X) to [out=180,in=45, looseness=1] (Y);
			\draw[->,>=latex] (Y) to [out=0,in=-135, looseness=1] (X);
			\draw[->,>=latex] (X) -- (Z);
			\draw[->,>=latex] (Y) to [out=-10,in=190, looseness=1] (Z);
			\end{tikzpicture}
			\\\midrule
			Dataset 10 &
			\begin{tikzpicture}[{black, circle, draw, inner sep=0}]
			\tikzset{nodes=minimum height=0.5cm,minimum width=0.5cm, font=\footnotesize}
			\node (X) at (1,0) {M.E} ;
			\node (Y) at (0,-0.5) {G.H.I} ;
			\node (Z) at (2,-0.5) {C.M.I};
			%				\draw[->,>=latex] (X) -- (Y);
			%				\draw[->,>=latex] (X) -- (Z);
			\end{tikzpicture} 
			&
			\begin{tikzpicture}[{black, circle, draw, inner sep=0}]
			\tikzset{nodes=minimum height=0.5cm,minimum width=0.5cm, font=\footnotesize}
			\node (X) at (1,0) {M.E} ;
			\node (Y) at (0,-0.5) {G.H.I} ;
			\node (Z) at (2,-0.5) {C.M.I};
			\draw[->,>=latex] (Y) -- (Z);
			%				\draw[->,>=latex] (X) -- (Z);
			\end{tikzpicture}
			& 
			\begin{tikzpicture}[{black, circle, draw, inner sep=0}]
			\tikzset{nodes=minimum height=0.5cm,minimum width=0.5cm, font=\footnotesize}
			\node (X) at (1,0) {M.E} ;
			\node (Y) at (0,-0.5) {G.H.I} ;
			\node (Z) at (2,-0.5) {C.M.I};
			\draw[->,>=latex] (X) to [out=180,in=45, looseness=1] (Y);
			\draw[->,>=latex] (Y) to [out=0,in=-135, looseness=1] (X);
			\draw[->,>=latex] (X) -- (Z);
			\end{tikzpicture}
			&
			\begin{tikzpicture}[{black, circle, draw, inner sep=0}]
			\tikzset{nodes=minimum height=0.5cm,minimum width=0.5cm, font=\footnotesize}
			\node (X) at (1,0) {M.E} ;
			\node (Y) at (0,-0.5) {G.H.I} ;
			\node (Z) at (2,-0.5) {C.M.I};
			\draw[->,>=latex] (X) to [out=180,in=45, looseness=1] (Y);
			\draw[->,>=latex] (Y) to [out=0,in=-135, looseness=1] (X);
			\end{tikzpicture}
			&
			\begin{tikzpicture}[{black, circle, draw, inner sep=0}]
			\tikzset{nodes=minimum height=0.5cm,minimum width=0.5cm, font=\footnotesize}
			\node (X) at (1,0) {M.E} ;
			\node (Y) at (0,-0.5) {G.H.I} ;
			\node (Z) at (2,-0.5) {C.M.I};
			\draw[->,>=latex] (X) to [out=180,in=45, looseness=1] (Y);
			\draw[->,>=latex] (Y) to [out=0,in=-135, looseness=1] (X);
			\draw[->,>=latex] (X) to [out=0,in=135, looseness=1] (Z);
			\draw[->,>=latex] (Z) to [out=180,in=-45, looseness=1] (X);
			\draw[->,>=latex] (Z) to [out=-145,in=-35, looseness=1] (Y);
			\draw[->,>=latex] (Y) to [out=-10,in=190, looseness=1] (Z);
			\end{tikzpicture}\\\bottomrule

		\end{tabularx}
	\end{adjustwidth}
\end{table}

\subsubsection{Complexity Analysis}
Here, we provide a complexity analysis on different constraint-based algorithms including our proposed method.

%Our proposed method benefits from a smaller number of tests compared to constraint-based methods that infer a window causal graph.
In the worst case, the complexity of PC in a window causal graph is bounded by $\frac{(d\gamma_{max})^2(d\gamma_{max} - 1)^{k-1}}{(k-1)!}$,
%%
%\begin{equation*}
%\frac{(d\gamma_{max})^2(d\gamma_{max} - 1)^{k-1}}{(k-1)!},
%\end{equation*}
%%
where $k$ represents the maximal degree of any vertex and $\gamma_{max}$ is the maximum number of lags. Each operation consists of conducting a significance test on a conditional independence measure.
Algorithms adapted to time series, as \texttt{PCMCI} \citep{Runge_2019}, rely on time information to reduce the number of tests.
\texttt{PCTMI} reduces the number of tests even more since it infers a summary graph. 
Indeed, \texttt{PCTMI}'s complexity in the worst case is bounded by $\frac{d^2(d - 1)^{k-1}}{(k-1)!}$.
%\begin{equation*}
%\frac{d^2(d - 1)^{k-1}}{(k-1)!}.
%\end{equation*}

Figure~\ref{fig:complex} provides an empirical illustration of the difference in the complexity of the two approaches on the three structures (v structure, fork, diamond), sorted according to their number of nodes, their maximal out-degree, and their maximal in-degree. The time is given in seconds. As one can note, \texttt{PCTMI} is always faster than \texttt{PCMCI}, the difference being more important when the structure to be inferred is complex.

\begin{figure}[H]
	%\centering
	%\begin{wrapfigure}{r}{0.53\textwidth}%[htb]%[ht!]
	%	\centering
	\begin{tikzpicture}[scale=0.85]
	\renewcommand{\axisdefaulttryminticks}{4}
	\pgfplotsset{every major grid/.append style={densely dashed}}
	\pgfplotsset{every axis legend/.append style={cells={anchor=west},fill=white, at={(0.02,0.98)}, anchor=north west}}
	\begin{axis}[
	xmin = 0.8,
	xmax = 3.2,
	%			xmode=log,
	log ticks with fixed point,
	xtick = {1,2,3}, 
	xticklabels = {V structure, Fork, Diamond},
	ymin=0,
	ymax=8000,
	yticklabels = {0, $0$, $2000$, $4000$, $6000$, $8000$},
	grid=minor,
	scaled ticks=true,
	ylabel = {time (s)},
	height = 4.5cm,
	width=8cm,
	legend style={nodes={scale=0.9, transform shape}}
	]
	\addplot[blue,smooth,mark=*, error bars/.cd, y dir=both,y explicit] plot coordinates{
		(1, 525.9998454093933) +- (219.3504390175089, 219.3504390175089)
		(2, 526.0872015714646) +- (275.74229599530514, 275.74229599530514)
		%		(3, 621.0207129478455) +- (143.14584096047648, 143.14584096047648)
		(3, 1414.8373223304748) +- (392.5530790235211, 392.5530790235211)
	};
	\addplot[brown,dashed,mark=star, error bars/.cd, y dir=both,y explicit] plot coordinates{
		(1, 3046.395077228546) +- (411.08618076051766, 411.08618076051766)
		(2, 3282.558140087128) +- (289.50136765471194, 289.50136765471194)
		%		(3, 5072.794156193733) +- (858.6444354985474, 858.6444354985474)
		(3, 7355.551018548012) +- (1872.5676169309697, 1872.5676169309697)
	};
	\legend{\texttt{PCTMI}, \texttt{PCMCI}}
	\end{axis}
	\end{tikzpicture}
	\caption{{Time} computation for \texttt{PCTMI} and \texttt{PCMCI}.}
	\label{fig:complex}%MDPI: Commas are only used for numbers with five or more digits. Please remove them in four-digit numbers, e.g., “1,200” should be “1200”. (Done)
\end{figure}
%\end{wrapfigure}

\subsubsection{{Hyperparameters' Analysis}}
{Here, we provide a hyperparameter analysis on CTMI with respect to the maximum lag $\gamma_{\max}$ and the k-nearest neighbor $k$. Using the same $10$ datasets that are compatible with the fork structure in Table~\ref{tab:structure} and generated as described in Section~\ref{subsec:sim_data}, we performed two experimentations. 
}

{First, we computed the average and the standard deviation of CTMI($X^2;X^3$) and CTMI($X^2;X^3 \mid X^1$) over the $10$ datasets while varying $\gamma_{\max}$ between $3$ and $10$. The results given in Figure~\ref{fig:hyper_lag_max_k}a show that CTMI is robust with respect to $\gamma_{\max}$ in the case of dependency and in the case of conditional independency: the mean of CTMI($X^2;X^3$) remains the same for $\gamma_{\max}=3,4,5$ and slightly decreases for $\gamma_{\max}=10$, and the standard deviation remains the same for $\gamma_{\max}=3,4$ and slightly increases for $\gamma_{\max}=5,10$. On the other hand, the mean and the standard deviation of CTMI($X^2;X^3 \mid X^1$) remain the same for all $\gamma_{\max}$.}

{Second, we again computed the average and the standard deviation of CTMI($X^2;X^3$) and CTMI($X^2;X^3 \mid X^1$) over the $10$ datasets, but now, while varying $k$ between $5$ and $100$. The results given in Figure~\ref{fig:hyper_lag_max_k}b show that CTMI is robust with respect to $k$ in the case of conditional independency, but suffers in the case of dependency when $k$ increases: the mean of CTMI($X^2;X^3$) slightly decreases between $k=5$ and $k=10$, but drastically decreases for $k=50$.
}

\begin{figure}[H]
	%\begin{wrapfigure}{r}{0.53\textwidth}%[htb]%[ht!]
	%	\centering
	{\captionsetup{position=bottom,justification=centering}\begin{subfigure}{.5\textwidth}
			\begin{tikzpicture}[scale=0.85]
			\renewcommand{\axisdefaulttryminticks}{4}
			\pgfplotsset{every major grid/.append style={densely dashed}}
			\pgfplotsset{every axis legend/.append style={cells={anchor=west},fill=white, at={(0.02,0.98)}, anchor=north west}}
			\begin{axis}[
			xmin = 0.8,
			xmax = 4.2,
			%			xmode=log,
			log ticks with fixed point,
			xtick = {1,2,3, 4}, 
			xticklabels = {3, 4, 5, 10},
			ymin=0,
			ymax=0.1,
			grid=minor,
			scaled ticks=true,
			xlabel = {$\gamma_{\max}$},
			height = 4.5cm,
			width=8cm,
			legend style={nodes={scale=0.8, transform shape}}
			]
			\addplot[blue,smooth,mark=*, error bars/.cd, y dir=both,y explicit] plot coordinates{
				(1, 0.04) +- (0.02, 0.02)
				(2, 0.04) +- (0.02, 0.02)
				(3, 0.04) +- (0.03, 0.03)
				(4, 0.03) +- (0.03, 0.03)
			};
			\addplot[blue,dashed,mark=*, error bars/.cd, y dir=both,y explicit] plot coordinates{
				(1, 0.011) +- (0.005, 0.005)
				(2, 0.008) +- (0.004, 0.004)
				(3, 0.009) +- (0.004, 0.004)
				(4, 0.008) +- (0.004, 0.004)
			};
			\legend{CTMI($X^2 X^3$), CTMI($X^2 X^3 \mid X^1 $)}
			\end{axis}
			\end{tikzpicture}
			\caption{}			
			\label{fig:hyper_lag_max_k_a}
		\end{subfigure}
		\begin{subfigure}{.5\textwidth}
			\begin{tikzpicture}[scale=0.85]
			\renewcommand{\axisdefaulttryminticks}{4}
			\pgfplotsset{every major grid/.append style={densely dashed}}
			\pgfplotsset{every axis legend/.append style={cells={anchor=west},fill=white, at={(0.02,0.98)}, anchor=north west}}
			\begin{axis}[
			xmin = 0.8,
			xmax = 4.2,
			%			xmode=log,
			log ticks with fixed point,
			xtick = {1,2,3, 4}, 
			xticklabels = {5, 10, 50, 100},
			ymin=0,
			ymax=0.1,
			grid=minor,
			scaled ticks=true,
			xlabel = {k},
			height = 4.5cm,
			width=8cm,
			legend style={nodes={scale=0.9, transform shape}}
			]
			\addplot[blue,smooth,mark=*, error bars/.cd, y dir=both,y explicit] plot coordinates{
				(1, 0.05) +- (0.04, 0.04)
				(2, 0.04) +- (0.03, 0.03)
				(3, 0.01) +- (0.006, 0.006)
				(4, 0.01) +- (0.006, 0.006)
			};
			\addplot[blue,dashed,mark=*, error bars/.cd, y dir=both,y explicit] plot coordinates{
				(1, 0.006) +- (0.009, 0.009)
				(2, 0.006) +- (0.007, 0.007)
				(3, 0.007) +- (0.003, 0.003)
				(4, 0.007) +- (0.003, 0.003)
			};
			\legend{CTMI($X^2 X^3$), CTMI($X^2 X^3 \mid X^1 $)}
			\end{axis}
			\end{tikzpicture}
			\caption{}			
			\label{fig:hyper_lag_max_k_b}
	\end{subfigure}}
	\caption{{{CTMI} with respect to the maximum lag $\gamma_{\max}$ in (\textbf{a}) and the k-nearest neighbor $k$ in (\textbf{b}) for dependent time series ($X^2$ and $X^3$ in the fork structure in Table~\ref{tab:structure}) and conditionally independent time series ($X^2$ and $X^3$ conditioned on $X^1$ in the fork structure in Table~\ref{tab:structure}).}}%MDPI: please add explanation for subfigure ab. (Done)
	\label{fig:hyper_lag_max_k}
\end{figure}

%%%%%%%%%%%%%%%%%%%%%%%%%%%%%%%%%%%%%%%%%%%%%%%%ù

\section{Discussion and Conclusions}\label{sec:chapEntropyBased_concl}
We addressed in this paper the problem of learning a summary causal graph on time series with equal or different sampling rates. To do so, we first proposed a new temporal mutual information measure defined on a window-based representation of time series. We then showed how this measure relates to an entropy reduction principle, which can be seen as a special case of the probability raising principle. We finally combined these two ingredients in PC-like and FCI-like algorithms to construct the summary causal graph. 
Our method proved to work well with small time complexity in comparison with similar approaches that use mutual information.
The main limitations of our methods are that 
they cannot orient causal relations that are not oriented by the classical PC-rules or FCI-rules and have a possible spurious correlation,
%is restricted to the Markov equivalence class in case of instantaneous relations 
and they assume acyclicity in summary causal graphs.

\bibliography{references}   % name your BibTeX data base

\end{document}